# A Survey on Self-Supervised Contrastive Learning for Multimodal Text-Image Analysis

**Asifullah Khan[1,2,3], Laiba Asmatullah[1,3], Anza Malik[1,3], Shahzaib[1,3] and Hamna Asif[1,3]**

**Affiliations:**

[1] Pattern Recognition Lab, DCIS, PIEAS, Nilore, Islamabad, 45650, Pakistan.

[2] Deep Learning Lab, Center for Mathematical Sciences, PIEAS, Nilore, Islamabad, 45650, Pakistan.

[3] PIEAS Artificial Intelligence Center (PAIC), PIEAS, Nilore, Islamabad, 45650, Pakistan.

## Abstract

Self-supervised learning is a machine learning approach that generates implicit labels by learning underlined patterns and extracting discriminative features from unlabeled data. without manual labeling. Contrastive learning introduces the concept of "positive" and "negative" samples, where positive pairs (e.g., variation of the same image/object) are brought together in the embedding space, and negative pairs (e.g., views from different images/objects) are pushed farther away. This methodology has shown significant improvements in image understanding and image text analysis without much reliance on labeled data. In this survey, we discuss the terminologies, categorization, recent developments, and applications of contrastive learning concerning text-image models. Specifically, we provide an overview of the approaches of contrastive learning in text-image models in recent years and categorize the approaches based on different model structures. Furthermore, we introduce and discuss the latest advances of the techniques used in the process, such as pretext tasks for both images and text, architectural structures, and key trends. Lastly, we discuss the recent state-of-the-art applications of self-supervised contrastive learning text-image-based models.

**Keywords:** Contrastive learning, Self-supervised learning, Text-Image analysis, Multimodal learning, cross-modal retrieval, vision-language models, CLIP, Data augmentations, ViT, MoCo

**Abbreviations**

| | |
|---|---|
| AI | Artificial Intelligence |
| ALIGN | Attention-based Language-Image Network |
| ARLMs | Autoregressive Language Models |
| BERT | Bidirectional Encoder Representations from Transformers |
| BYOL | Bootstrap Your Own Latent |
| ChiLS | Child-Learner Systems |
| CL | Contrastive Learning |
| CLIP | Contrastive Language–Image Pre-training |
| CLOOB | Contrastive Leave-One-Out Benchmark |
| CL-VQA | Contrastive Learning for Visual Question Answering |
| CNN | Convolutional Neural Network |
| CoCa | Contrastive Captioner |
| COOKIE | Contrastive Open-World Knowledge Integration |
| CPC | Contrastive Predictive Coding |
| DCL | Decoupled Contrastive Learning |
| GPT-3 | Generative Pre-Trained Transformer 3 |
| InfoLOOB | Information Leave-One-Out Bound |
| InfoNCE | Information Noise Contrastive Estimation |
| LIMoE | Learned Input Mixture of Experts |
| MedVQA | Medical Visual Question Answering |
| MixCon3D | Mixed Contrastive 3D |
| MLM | Masked Language Modeling |
| MoCo | Momentum Contrast |
| MoE | Mixture of Experts |
| MosaiCLIP | Mosaic Contrastive Language–Image Pre-training |
| NLP | Natural Language Processing |
| NSP | Next Sentence Prediction |
| NSO | Noise-Contrastive Estimation for Structured Outputs |
| NT-Xent | Normalized Temperature-scaled Cross Entropy |
| PubMedCLIP | PubMed Contrastive Language–Image Pre-training |
| RA-CLIP | Region-Aware Contrastive Language–Image Pre-Training |
| ResNet | Residual Network |
| SimCLR | Simple Contrastive Learning of Representation |
| SOP | Sentence Order Prediction |
| SSL | Self-Supervised Learning |
| TCN | Temporal Convolutional Network |
| TCL | Temporal Contrastive Learning |
| VideoCoCa | Video Contrastive Captioner |
| ViT | Vision Transformer |
| VizWiz | Visual Question Answering for Blind Users |
| VQA | Visual Question Answering |
| ZSL | Zero-Shot Learning |

# 1. Introduction

The evolution of powerful hardware, the availability of large annotated datasets, and powerful deep learning techniques (Khan et al., 2020) have significantly accelerated progress in machine learning. Deep learning heavily relies on computational power and vast amounts of data for improving its generalization capabilities (Goel et al., 2020). Large language models, for instance, are data hungry, need large datasets to train on, and require significant computational power to learn meaningful representations. Tasks such as Image captioning, language translation, and cross-modal retrieval heavily rely on AI-specialized hardware (GPUs and TPUs) and large curated datasets. Hardware is evolving at a rapid pace, however high-quality data still faces challenges.

The importance of self-supervised learning becomes clear when we note how traditional machine learning techniques struggle to address limitations due to reliance on annotated datasets. The process of manually annotating large and immense datasets is often expensive, time-consuming, labor intensive, and requires a domain expert for specialized tasks. At the same time growth of unlabeled digital data expands at a rapid pace due to our heavy reliance on social media and digital technologies. This highlights the growing need for SSL techniques to handle extensive amounts of data, to reduce dependency on manual annotation, all while saving time and money (Khan et al.,2024). In the context of deep learning, reducing reliance on an annotated datasets not only reduces cost, but also improves accuracy and robustness. To achieve this, Self-supervised contrastive learning is employed. Multimodal data is used to learn relationships between two or more modalities using paired instances. Self-supervised learning method learns complex patterns and relationships from unlabeled data by training models to distinguish similar pairs from dissimilar ones, without explicit labels.

Large-scale annotated datasets such as ImageNet (Deng et al., 2009) and COCO (Lin et al., 2014) show impressive results using supervised learning tasks, but still face challenges in certain tasks. This is because some applications require rare and highly specialized annotated data, which is too expensive to be annotated (Gomez et al., 2019)., while the modern specialized contrastive learning approaches not only learn representations from ImageNet, but also exceed accuracy and prove greater robustness in downstream tasks than supervised learning approaches (Khosla et al., 2021). Self-supervised learning approaches do more than just improving accuracy. They are also capable of performing well in disadvantageous conditions like noisy data and input corruptions (Zhong et al., 2022).

This paper explores advancement and challenges in self-supervised contrastive learning, particularly focusing on pretext tasks such as jigsaw puzzle (Misra & Maaten, 2019), geometric transformation, and viewpoint predictions (J. Wang et al., 2024). These pretext tasks significantly improve models' capabilities to capture local and global features effectively and to learn robust feature representations from an unlabeled dataset. The survey also discusses visual and contextual domains. It explores textual techniques like Masked language modeling (MLM) (Kwon et al., 2023), next sentence predictions (NSP), and sentence permutation. These help grasp complete contextual relationship coherence. The incorporation involves different sensory modalities and improves cross-modality alignment. It includes use of methodologies like TCN (Sermanet et al., 2018), MoCo (He et al., 2020), and the CLIP technique. CLIP (Radford et al., 2021) achieved remarkable fame in generative AI by processing text-image pair data, enabling it to learn from multimodal representations. CLIP, based on aligning textual and visual information in a shared embedding space, shows impressive capabilities across various vision and language tasks. Mean, while, the idea of aligning different modalities in a shared embedding space revolutionized the field of cross-modality.

Contrastive learning and SSL merges with supervised learning using hybrid loss functions. This tackles labeling challenges in large datasets. This improves robustness and model generalization across domains in real-time applications. The text-image models use cross-modal contrastive learning. Concept of matching text and visual representations, while preserving modality-specific properties. To improve model generalization, approaches including zero-shot learning (ZSL), hierarchical categorization, linear probing, and transfer learning are employed to adapt to different tasks. These techniques prove versatility across image classification, text-image models, visual Q&A, 3D vision and others. They bridge the gap between unsupervised and supervised learning.

This paper overall covers self-supervised contrastive learning approaches in text and image analysis, different pretext tasks for images and texts, an architectural view of text-image-based multimodal, challenges, and integration of self-supervised learning into supervised learning tasks.

The remaining paper is organized as follows. Section 2 presents the Background of SelfSupervised learning, Section 3 discusses pretext tasks for images and texts, Section 4 focuses on Contrastive learning approaches in text-image models, and key trends in Contrastive approaches., while Section 5 explores an architecture overview of contrastive learning, TextImage-based models. Section 6 describes integration of Self-Supervised learning into supervised learning tasks, Section 7 explains challenges with integration and their solutions, and Section 8 highlights the application of contrastive-based text-image analysis. Finally, Section 9 concluded the survey paper.

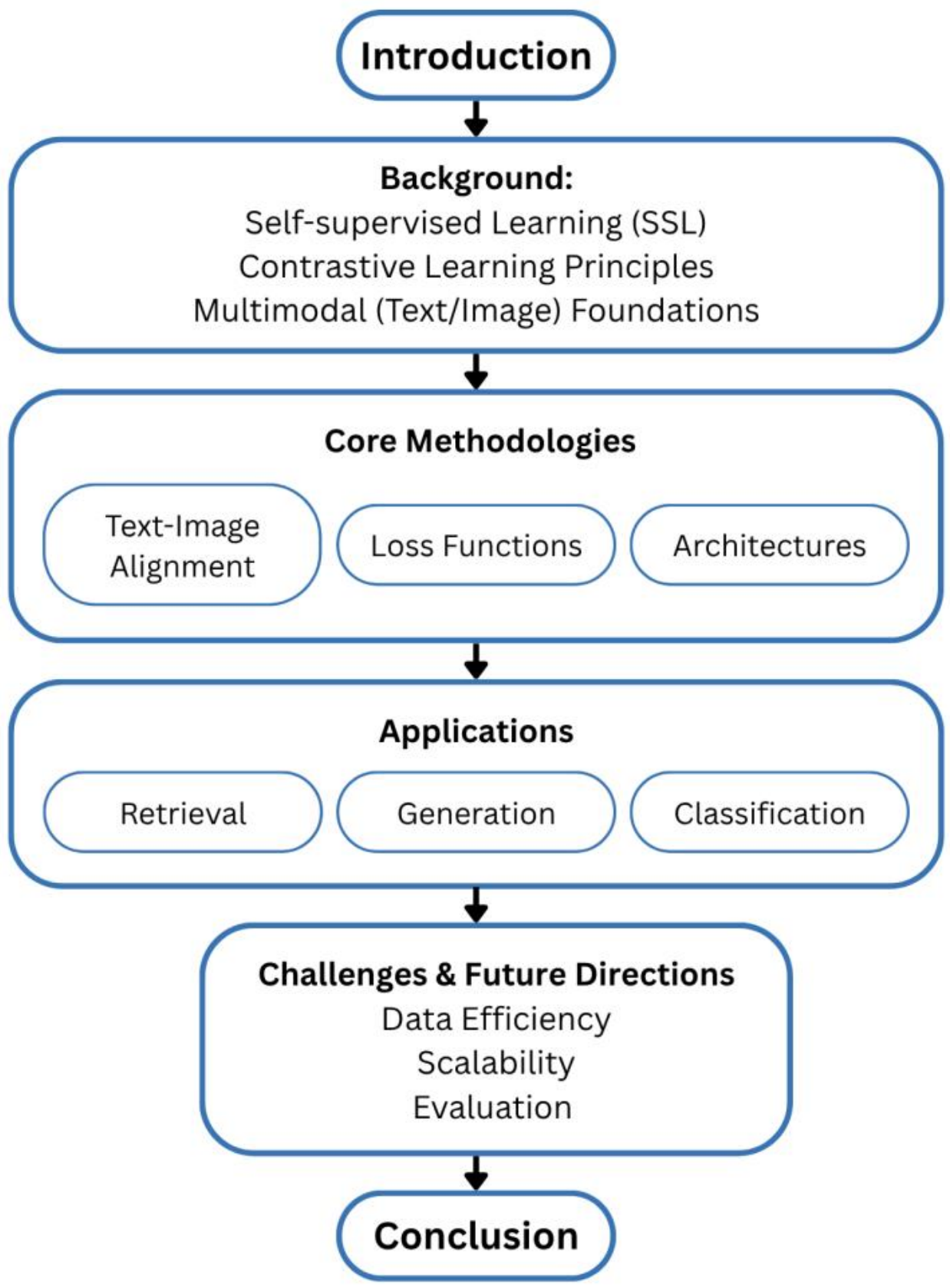

**Figure 1** Road Map

# 2. Background

## 2.1 Unimodality

Unimodality refers to the process of dealing with a single type of data. The unimodality AI models are tailored to work with a single type of data source. Those models are trained to execute a specific task, while being highly specialized and optimized within a particular domain. The model's simplicity within a specific domain allows its efficient training, clear understanding, and deep specialization within a focused task. It learns from the raw features of the modality (i.e. text, image, audio, video, etc.),, but faces struggles in exposure to diverse, complex multiple data types and changing domains. It is restricted to a single application, specialized task, and focused domain. unimodality faces struggles in limited data processing and broader context capabilities.

## 2.2. Multimodality

Multimodality refers to the process of learning representation from different types of input data. These types of models address the limitations of dependency on a single and specific task. The multimodality AI models use a variety of data sources, including text, images, audio, and video, to provide more context and a comprehensive understanding. It combines multiple data sources and integrates them into a common model capable of handling multiple data types. The model's complexity increases with different modalities and integrations,, but it helps models in understanding broader context and big data processing. The fusion of multiple modalities, such as image and text fused corpus, enhances model prediction and interpretability, allowing it to deal with real-world situations in the same manner that humans do.

## 2.3 Self-Supervised Learning

Unfortunately, there is a large amount of unlabeled data in the actual world, and machine learning models face challenges to learn from this data. To overcome the limitations of supervised learning (relying on labeled data), various other pattern learning approaches have emerged in deep learning, including active learning, semi-supervised learning, and SelfSupervised learning. The need for self-supervised learning arises when supervised learning techniques encounter challenges, such as reliance on manually labeled datasets. Therefore, SSL techniques gained attention as an alternative method due to their data efficiency and generalization capabilities (Ruslim et al., 2023).

Self-Supervised Learning is a branch of unsupervised learning that eliminates the concept of manual labeling. It learns to discriminate features from unlabeled data, removing reliance on human annotations. It stands out as a transformative approach for utilizing unlabeled data by introducing pseudo-labels via self-defined pretext tasks such as predicting image rotations, reconstructing shuffled patches, or filling missing words in a sentence. It extracts useful features from data by solving a *pretext task* before tackling the main task. A pretext task is a placeholder problem that doesn't require labeled data, but helps the model understand patterns within the data. The Self-Supervised learning models learn meaningful representations from massive amounts of unlabeled data, resulting in better feature extraction and performance in downstream supervised tasks, making SSL a bridge between unsupervised and supervised learning.

## 2.4 Contrastive Learning

Contrastive learning has greatly advanced the field of self-supervised learning by focusing on relationships between similar and dissimilar data instances rather than label-based data points. The concept has drawn attention in self-supervised learning due to the efficient utilization of huge amounts of unlabeled data, particularly in applications such as image and textual representations.

Contrastive learning uses the idea of 'positive' and 'negative' samples. Positive pairs (like variations of the same image) are brought closer in the embedding space, while negative pairs (from different images) are pushed apart. The model learns by bringing similar pairs together and pushing dissimilar pairs farther apart. The objective is to minimize the distance between semantically similar pairings in the feature space, while maximizing the distance between dissimilar pairs. Augmentations in contrastive learning provide different and useful training samples. It creates positive pairs by various transformations like rotation, cropping, and geometrical alterations. The model learns by comparing samples, adhering to the SSL principle of keeping variational samples from the same class close together in the embedding space and separating those from other classes.

SIMCLR (Chen et al., 2020) is a neural network encoder used in contrastive learning to convert pictures into feature vectors. It’s a widely recognized framework because it uses data augmentations to create positive/negative pairs, helping models distinguish between augmented and different images. The technique used the InfoNCE loss function for optimization and learning representations from positive and negative samples. BYOL (Grill et al., 2020), on the other hand, is a newer learning technique. It evolves in contrastive learning by achieving high-quality visual representations without requiring negative samples. BYOL uses two neural networks, an online and a target network,, while learning from each other during interaction. It follows a teacher-student training mechanism. An online networks learn meaningful representation from augmented images, while target networks provide stable representation for online networks to match. The loss is computed using the similarity between the online network and the latent representation of the target network. The loss decreases as the similarity between the representation of online and target neural networks increases. Unlike traditional contrastive learning, BYOL doesn’t rely on comparing positive and negative pairs like SimCLR and MoCo. Instead, it enhances similarity between two variational views of the same image using the coordinated learning process of the online and target networks (Zheng et al., 2021). The SSL approaches learn representations from an unlabeled dataset by solving auxiliary and automated tasks, encouraging models to understand contexts and underlying data structures. The model is trained on different auxiliary tasks like transformative, order-based based and rearrangement tasks to capture local and global features. A well-designed pretext task helps the model improve generalization capabilities and learn feature semantics. After training on automated tasks, the model is fine-tuned for the specific tasks like classification, detection, and recognition as per need (Ruslim et al., 2023).

# 3. Pretext Tasks

Pretext task is an auxiliary and automated task in SSL to train a model, learn underlying features, capture semantic meaning, and extract meaningful representations from unlabeled data before fine-tuning the model to a specific purpose. The task acts as a mechanism for training a model on unlabeled data. The task itself is a problem that the model tries to solve using logic to understand underlying patterns. The basic purpose of the pretext task is to guide the model toward learning features that generalize well. Tasks such as predicting a missing word in a sentence during training guide a model to learn the relationship and contextual meaning among words. Training a model to solve an automated task is not the primary objective of the model, but instead focuses on building foundational understanding about the structure of data.

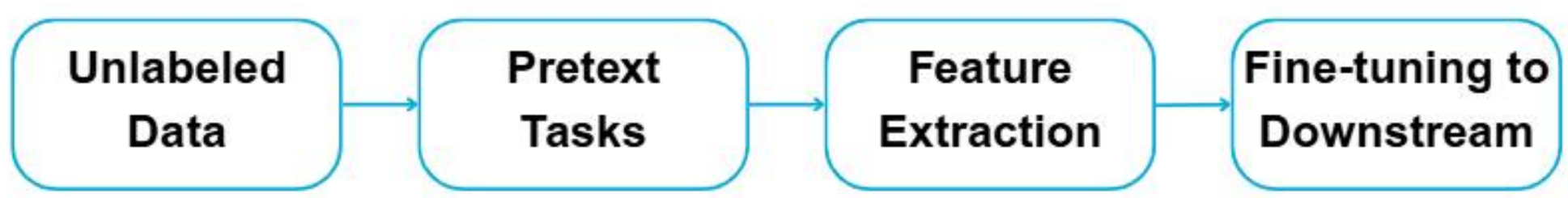

**Figure 2.** Flow Diagram representing the general process for Pretext Tasks

## 3.1. Pretext Tasks for Images

Image pretext tasks can be categorized into two primary configurations: relational-based approaches and transformational-based approaches. Relation-based methods, such as MoCo (Momentum Contrast) and SimCLR, differentiate positive and negative samples. Transformation-based techniques, on the other hand, utilize data augmentation to generate new classes. For example, these methods evaluate how visuals are relative to one another in terms of spatial properties, completing jigsaw puzzles, determining how much images are rotated, as well as correctly ordering reshuffled patches.

Pretext tasks are meant to drive the network to learn things like edges, shapes, colors, and textures that will be useful in later problems. The tasks help to define training objectives for representation learning and are a key component of self-supervised learning.

### 3.1.1. Color Transformation

Color transformation is the process of modifying an image's color characteristics, while preserving its base context intact. It creates false labels to diversify the dataset. This pretext task applies various random alterations like lighting, contrast, or color on images. Image transformations include random variations in brightness, contrast, saturation, and hue, called color jittering. Additionally, techniques such as gray scaling, histogram equalization, channel dropping, and color normalization are also categorized in color transformation. The proposed pretext techniques push the model to obtain invariant representations across different color distri, butions, leading it to learn features that generalize well across multiple tasks.

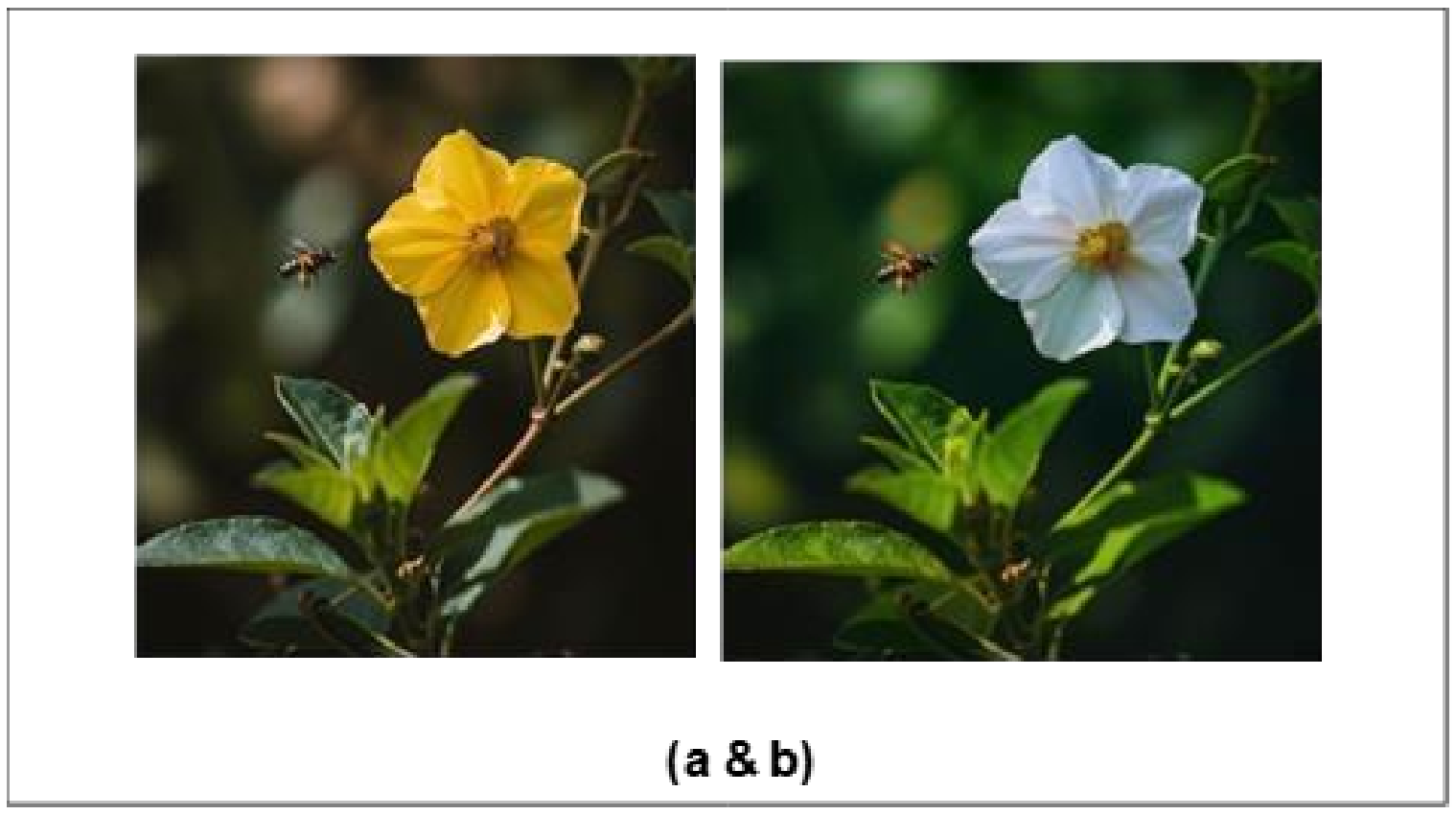

**Figure 3.** Color Transformation serves as a pretext task. (a) Original. (b) Transformed image using Color Jitter with parameters adjusted to completely transform image color by selecting an optimal value of saturation and hue.

### 3.1.2. Geometric Transformation

Geometric Transformation is a data augmentation approach that allows self-supervised learning models to be invariant to changes in the location, orientation, or spatial properties of the components of an image. It does this by appending variances through operations such as performing random cropping, re-sizing, scaling, shifting, and horizontal or vertical flipping at a defined angle. These changes alter the spatial configuration of the image, but preserve its semantic information, so the model can see different viewpoints of the same object. By training the model on tasks like rotating photographs at specific angles, it can learn invariant features from different augmented images and help in improving its generalization capabilities for unseen data. The learned features from augmented images enable the model to predict and interpret images with variations both in orientation and position, making it far more robust and useful than a series of images, again without the need for any input from the human.

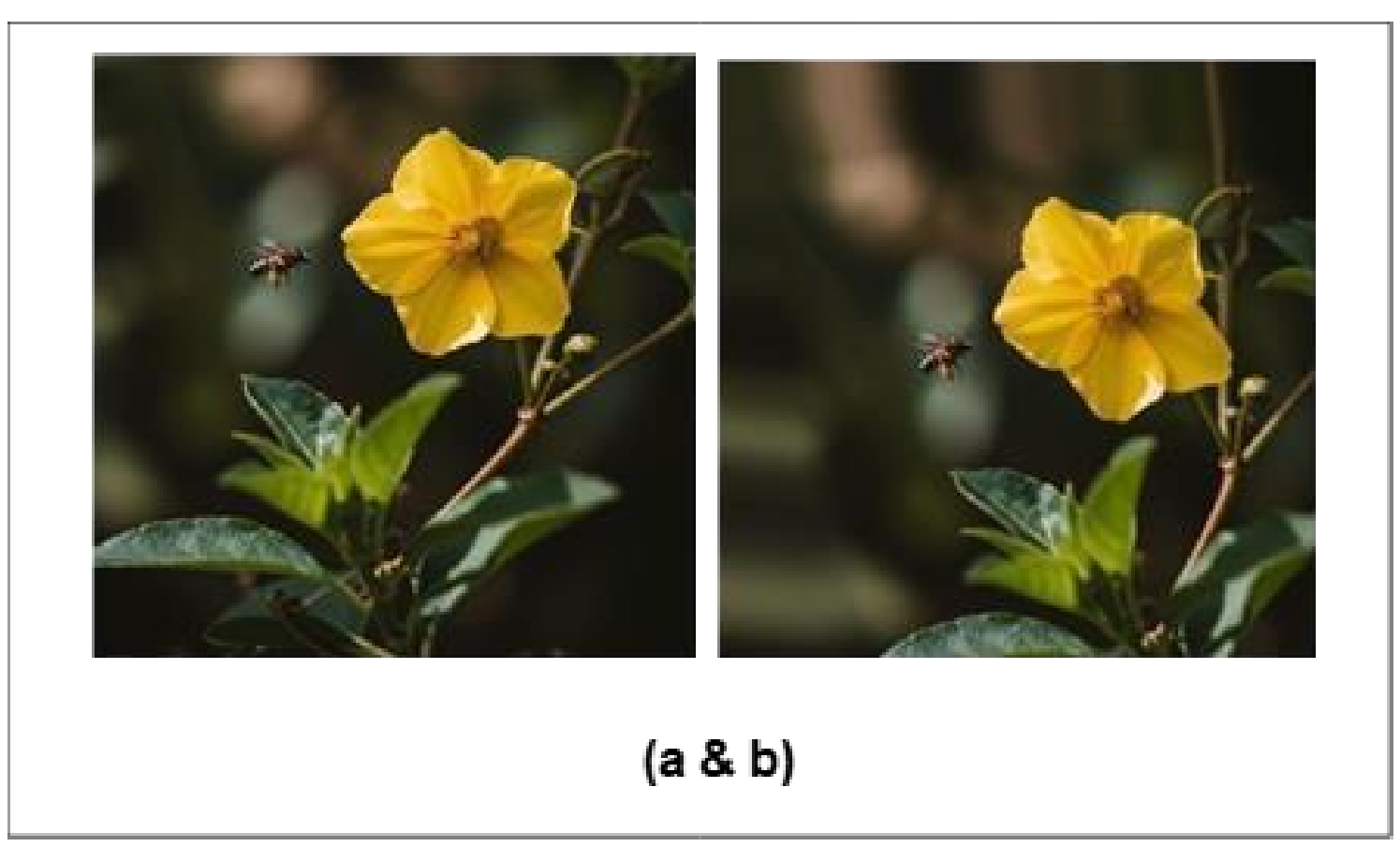

**Figure 4.** Geometric Transformation serves as a pretext task. (a) Original. (b) Transformed image using a combination of shifting, shearing, and changing angle. Adjustments include angle=15, shift-x=30, shifty=30 and shear-factor=0.1.

### 3.1.3. Gaussian Blur

Gaussian Blur is a strong pretext task, influences & uses smoothing filters that reduce image noise and detail by averaging the pixel values within a neighborhood, weighted by a Gaussian function. The Gaussian function assigns greater weight to pixels closer to the neighborhood's center, and gradually less weight to those further out. Gaussian blur is particularly suitable to context-based pretext tasks as it clears fine-grained features and exposes broader areas. By incorporating Gaussian blur as a pretext task, the model can effectively handle fluctuations in image quality. By learning different variant features from Gaussian blurred images, the model learns to recognize important and invariant properties and is more robust and generalizable for subsequent tasks like classification and object detection, even if the input data is not clear.

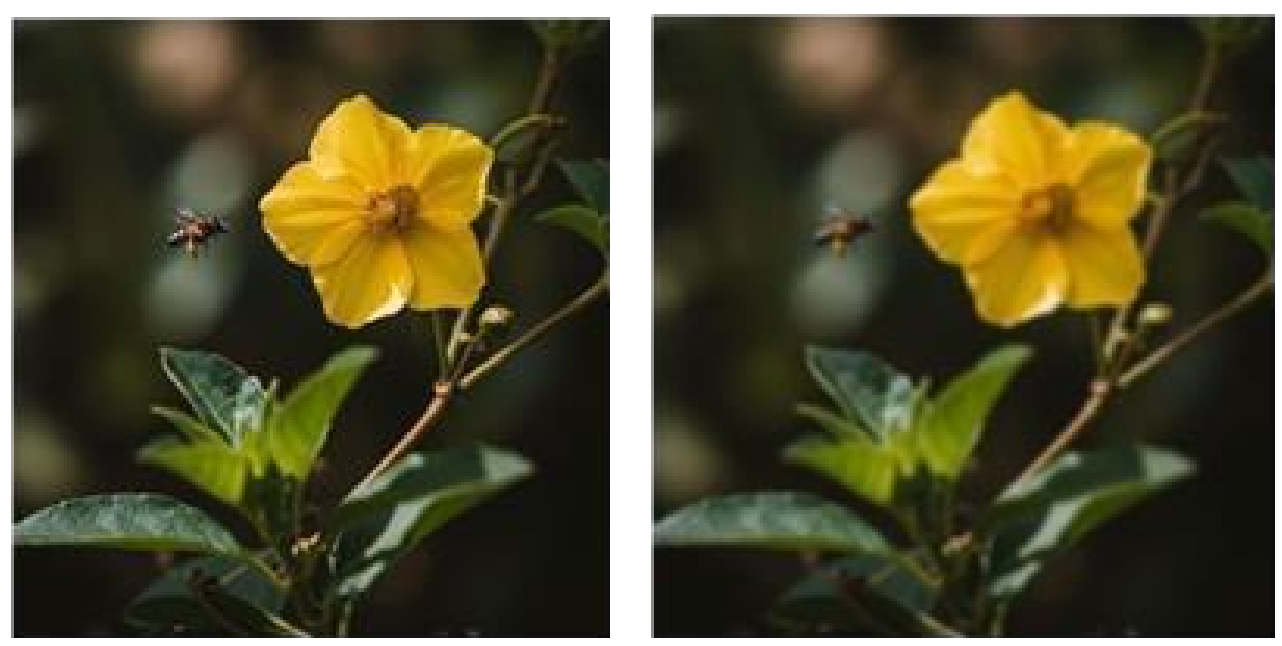

(a & b)

**Figure 5.** Gaussian Blur serves as a pretext task. (a) Original. (b) Transformed image with Gaussian blur applied using a kernel size of 7*7 and a sigma range of 5.0 to 10.0 with smoothing filtering.

### 3.1.4. Instant Discrimination

Instant Discrimination is an advanced pretext task where the model learns downstream representations by distinguishing between a variational view of the same image and completely different image pairs. It does it by applying color transformation as pretext tasks (Gaussian blur, Gaussian noise, color distortion, and grayscale transformation). These transformations add variety to the dataset and force the model to pay attention to learning meaningful features. Apart from color changes, spatial augmentations like cropping, re-sizing, flipping, and scaling are applied. These augmentations allow models to learn basic feature representations.

Instance discrimination used Contrastive loss function, since it is responsible to learn representations during pretext training. It aligns positive features (those from multiple enhanced perspectives of the same image), while increasing the separation of negative features (those from distinct photos). This method enables the model to develop robust and transferable representations that are useful for downstream tasks (Lu, 2022).

### 3.1.5. Global-Local Contrast

The purpose of this type of pretext task is to learn the lower-level, more fine-grained, and spatial representations that can effectively be linked to the global context of an image.

#### 3.1.5.1. Jigsaw Puzzle

Jigsaw Puzzle (Noroozi & Favaro, 2017) is a self-supervised learning method used during model training to learn both local and global features from images without requiring labels. In this approach, the original image serves as the base or anchor. The image is divided into multiple patches based on a specified grid size. These patches are then randomly shuffled to introduce randomness, and the shuffled patches act as positive samples for the anchor image. The model applies mathematical functions to reconstruct the original image by finding the correct positions of the shuffled patches. During estimating the correct patches, the model learns local and global features of patches to rearrange them in a correct order. This auxiliary task is designed to understand the broad picture of an image. The method is based on context learning, grasping detailed information, and detecting discriminative features.

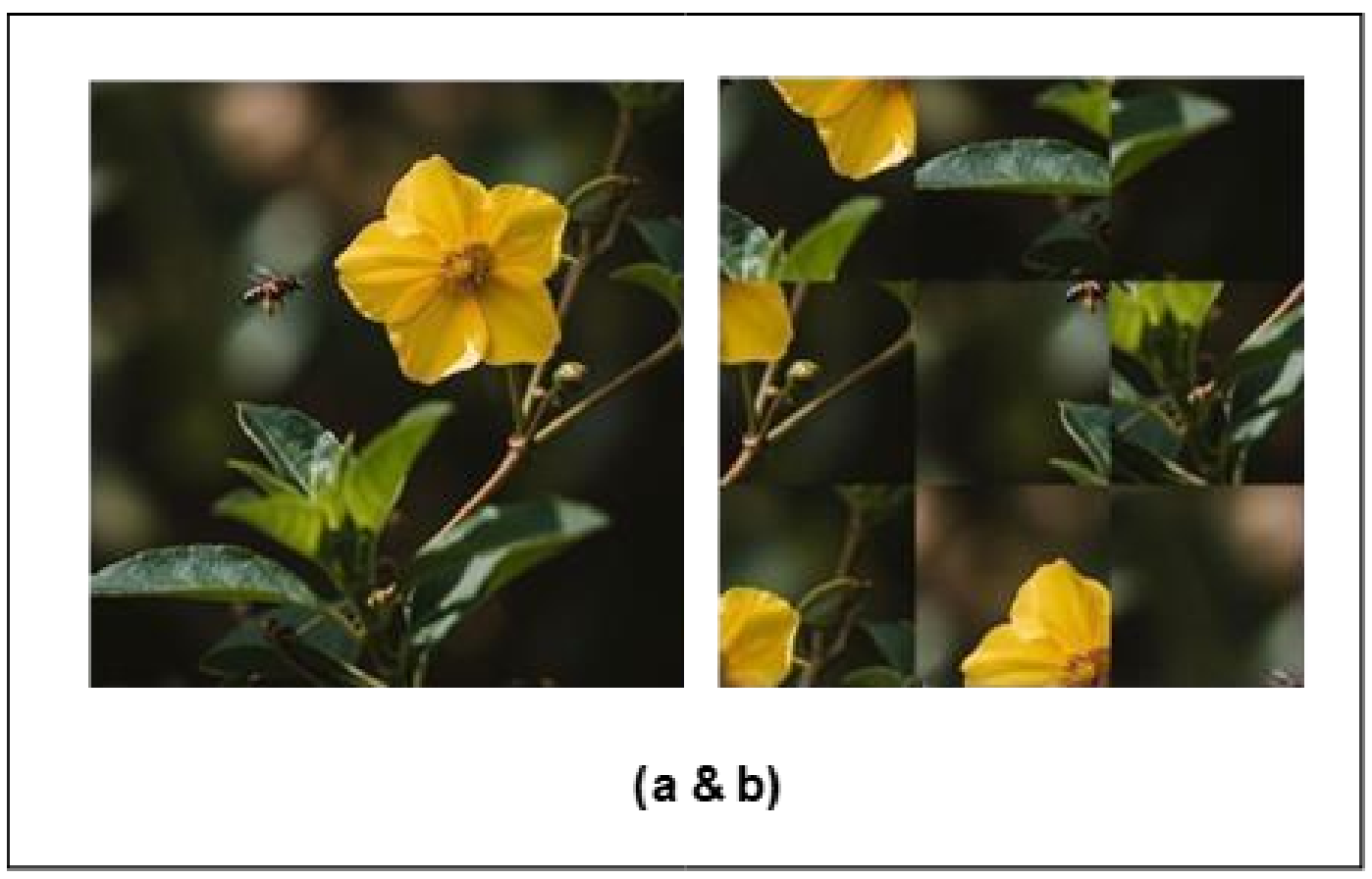

**Figure 6.** A jigsaw puzzle serves as a pretext task. (a) Original. (b) Jigsaw puzzle created by dividing the original image into a 3x3 grid of patches, shuffling the patches, and reassembling them.

#### 3.1.5.2. Jigsaw ViT

**Jigsaw ViT** (Y. Chen et al., 2023) is inspired by the jigsaw puzzle pretext task., but this auxiliary task is employed in transformer models for various vision-based tasks, like recognition and verification tasks. The basic purpose of this task is to enhance model generalization capabilities in image recognition. This method utilized the same concept of reconstructing the original image from shuffled patches to predict the correct order., but jigsaw ViT is more complex than the standard jigsaw puzzle task. It is because the positional embedding in the jigsaw puzzle task, the patches are placed concerning one another in the image., but Jigsaw ViT alters this early on: it removes positional embedding altogether so that the model is not dependent on the fixed positions of the image patch. For this reason, in contrast to CNNs, the ViT architecture does not require positional embedding (Khan et al., 2024) as it can learn to recognize the spatial relationships between patches without direct manipulations, thus enabling a richer and more flexible understanding of the image geometry. Randomly removing some patches from the image during training is another key feature in Jigsaw ViT. This ViT model is trained to understand the spatial relationship between patches using a self-supervised way, and to predict the correct arrangement of patches., while this pretext task does not rely on positional embedding, which helps the model to understand contextual and global clues instead of the spatial relationship between patches.

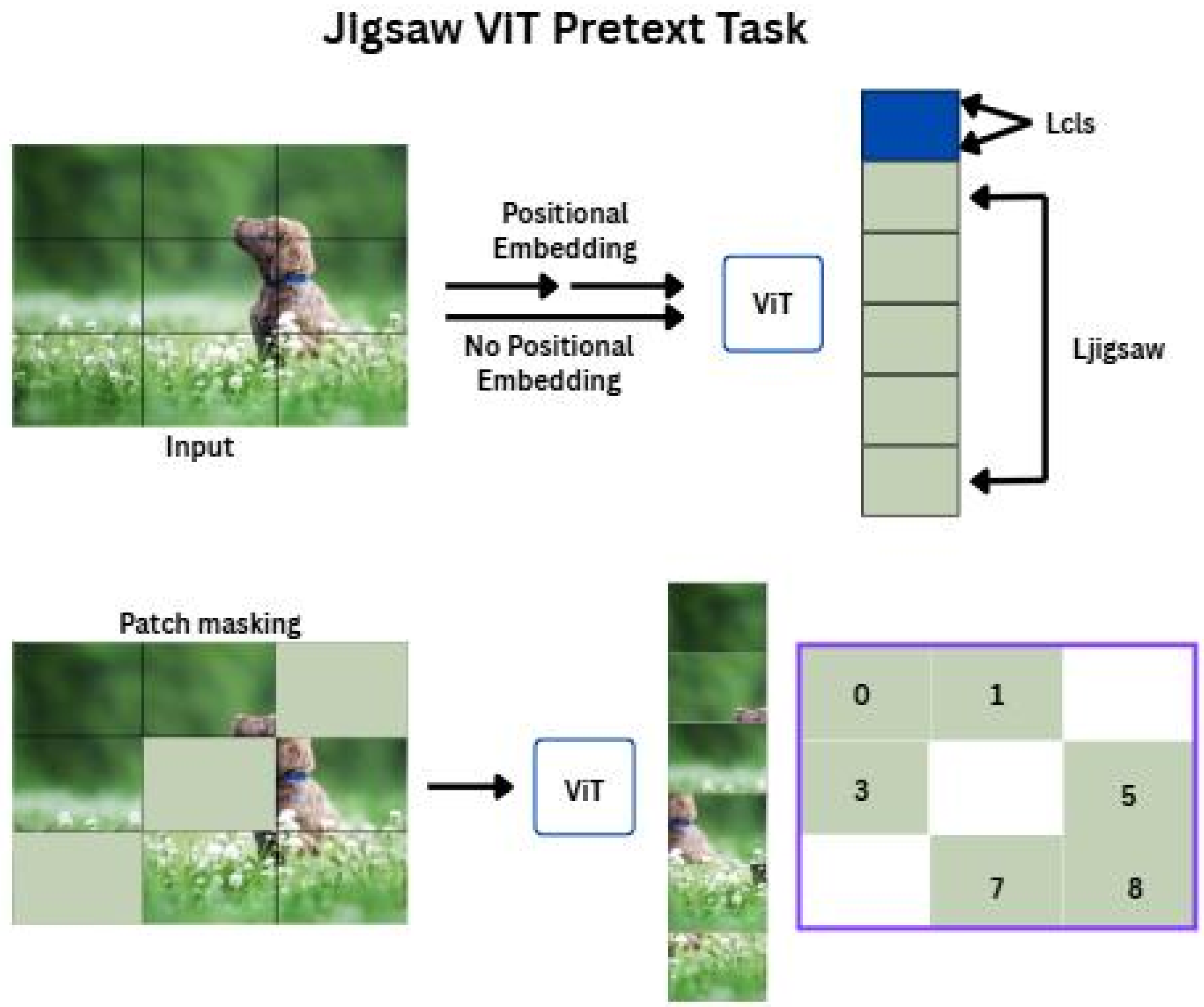

**Figure 7.** Illustrates Jigsaw **ViT** pretext tasks for vision in transformers, where images are shuffled and reordered to enhance the spatial relationship between patches. The pink cell represents the traditional ViT task for classification by adding positional embedding to maintain spatial relationships. The new blue cells represent a self-supervised learning task, to mask some patches, remove positional embedding, and to predict the correct position of unmasked patches. The L(jigsaw) shows loss to the optimized model, which helps in learning the spatial relationship between patches.

### 3.1.5.3. Frame Order Based

This Pretext Task will be suitable for temporally varied data, such as sequential sensor data (e.g., real-time MRI) or video sequences. This task allows for the encoding of temporal signals, which enables us to learn useful patterns through sequential data. Conceptually, it is like the jigsaw puzzle task,, but it allows you to reshuffle the frames of the same video/clip according to their time intervals. In this approach, the frames of the same video are treated as positive instances,, while frames from other videos in the dataset are treated as negative instances. The model learns representations by identifying the temporal relationships between sequences of frames. It learns to predict the correct order of frames that are relatively closer in terms of their timestamp as well as semantic continuity. The causally related frames, temporally linked to each other, and where there is an event progression, are arranged in a correct order. The model learns during training to arrange the shuffle frames in the correct order, while maintaining event progression with time. This enables models to combine frames of the same video to give the correct sequence due to capturing local and global temporal features. The model understands semantic relationships between frames.

#### 3.1.5.4. Contrastive Predictive Coding

CPC (Oord et al., 2019) is an unsupervised learning pretext task that generates future data points from the past context. The model processes input data, mapping it into lower lowerdimensional space called the latent space. Then, a trained separate auto regressor processes the lower-dimensional data and builds its context. The context is the summary of the data, which shows what happened so far. Instead of reconstructing the original input like a simple autoencoder, CPC compares the real future value of data with random ones to help the model learn from it. The loss function is being used to show how future data relates to our current context. The task is being used in images, audio, or sequential data using encoders like GRU, LSTM, and the transformer model.

#### 3.1.5.5. Viewpoint Prediction

Viewpoint prediction is an auxiliary task involved during training to predict the viewpoint of the camera. The model is trained on huge videos captured from multiple angles. The technique used the SSL idea to eliminate the human effort to label data,, but instead focuses on learning similarities and differences in videos taken from different viewpoints. The model uses the concept of contrastive learning, where similar frames having varying angle orientation are placed closer to each other,, while frames that depict different actions or temporal sequence are further apart in embedding pace. This pretext task ignores distractions like lighting variations, background noise, and other disturbances., but instead, it focuses on key features of objects like joint movement, different poses, and attributes of objects like human poses and human-object interactions. This automated task applies Temporal Convolutional Networks (TCNs) to develop representations that remain consistent across varying viewpoints by comparing visually similar frames over different times and angles, and improves the model's ability to perceive and replicate human actions.

## 3.2. Pretext Tasks for Text

Pretext Tasks for Images is fundamental for the application of tasks such as translation and sentiment analysis, where labeled data might be scarce. The tasks generate supervisory signals from the data itself, allowing for the extraction of latent features and avoiding the preprocessing of manually annotated data. Training through pretext tasks reduces Overfitting by exposing the models to diverse linguistic patterns and structures during training facilitates Transfer Learning since the representations learned through pretext tasks form an ideal foundation for enabling models to get fine-tuned efficiently on special tasks with limited labeled data and thoroughly improve performance on downstream tasks in general. NLP includes pretext tasks like Masked Language Modeling (MLM), Next Sentence Prediction (NSP), Sentence Order Prediction (SOP), Autoregressive language modeling, and Sentence Permutation.

### 3.2.1. Masked Language Modeling (MLM)

MLM masks words in a sentence and trains the model to predict them based on the surrounding context, helping the model to understand word relationships and context.

The masked image contrastive learning model would propose a different framework in terms of masked tokens, generating contrasting views of images learned by the high-level semantic feature through contrastive learning., while randomly masking portions of an image, MiCL creates different types of views on the same instance that highlight and emphasize fine-grained conceptual differences, such that visual representation learning is increased without requiring augmentation data or auxiliaries, as suggested by (*Masked Image Contrastive Learning for Efficient Visual Conceptual Pre-Training*, n.d.).

ConLIP (Luo et al., 2022) is another framework that aligns images and texts in an instance-level representation that will have MLM for token-level alignment. Thus, it conditions masked text tokens based on the corresponding image representations toward facilitating cross-modal retrieval tasks. This approach leverages both instance-level and token-level interaction toward improving the model's performance under various image-text dense retrieval scenarios.

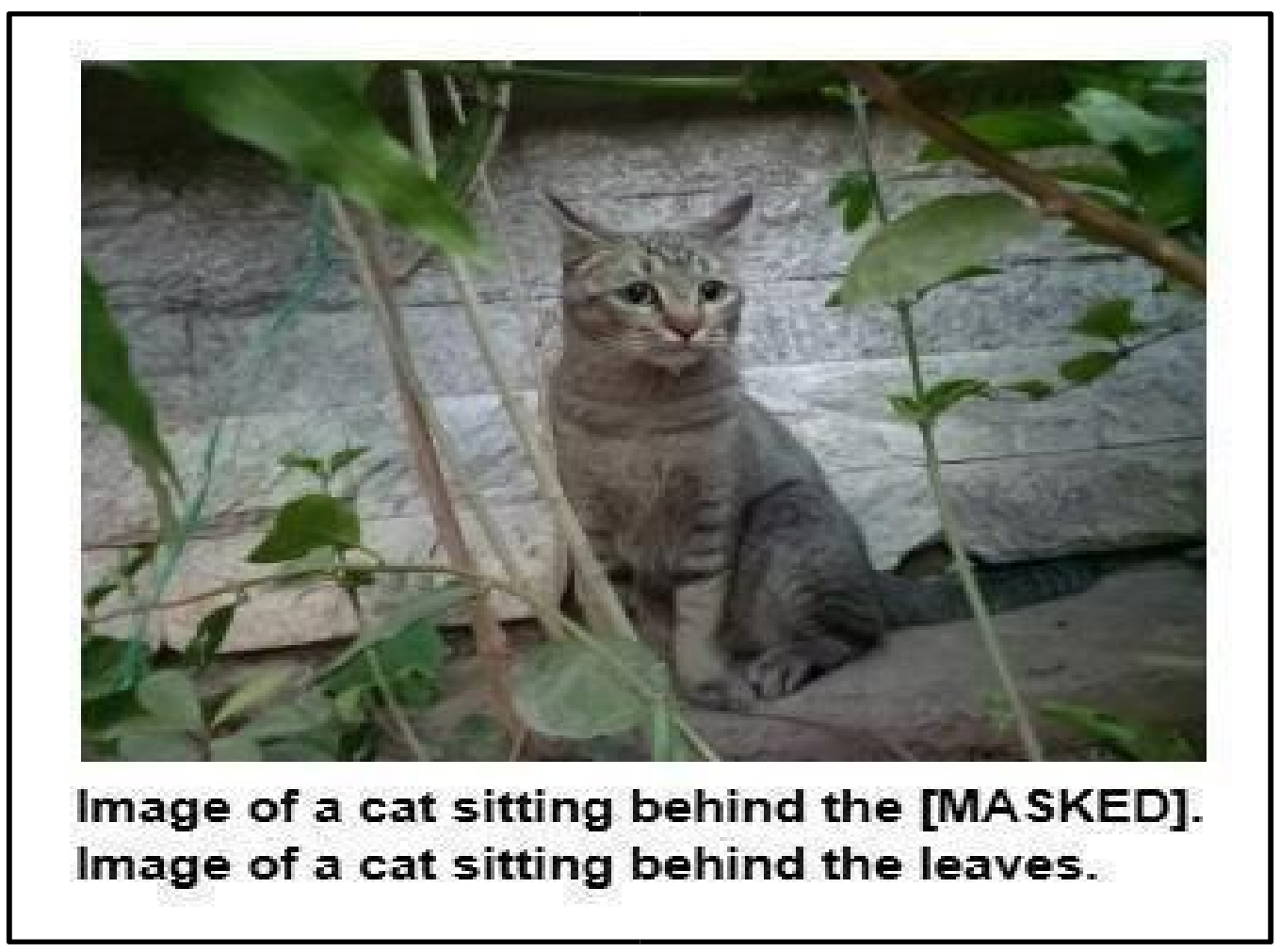

**Figure 8.** Masked Language Modeling as a Pretext Task for Text.

### 3.2.2. Next Sentence Prediction (NSP)

NSP improves the model's understanding of the relationship between the sentences. This is a critical task for natural language inference and question answering, and is done via the model being trained to predict whether one sentence logically follows another.

The Contrastive Learning-based Multi-Modal Architecture for the Emoticon Prediction model utilizes a dual-branch encoder architecture. It combines contrastive learning with NSP to map text and images into a common latent space and is based on understanding the relationships among sentences, visuals, and emoticons by analyzing how emoticons enhance textual content. The joint training approach exploits NSP to enhance the ability of the model to predict emoticons in a better manner, given both text and image contexts, than any other existing multimodal approaches (Pandey & Vishwakarma, 2024).

(Cao et al., 2022) presents a momentum contrastive learning model designed for sentence embedding, which can be extended to image-text scenarios. The role of negative samples in enhancing the performance of learned representations is discussed; the length of the negative sample queue can have a significant impact on outcomes in text contrastive learning models. This approach suggests possible applications in multimodal settings where images and texts are analyzed together.

### 3.2.3. Sentence Order Prediction (SOP):

SOP is a task that involves establishing whether two sentences are in the correct chronological order or not. To learn about the narrative structure and coherence, and flow of information, NLP models have to acquire knowledge, pretext tasks like SOP enhance the capacity of the model to understand what sentences relate to each other based on contextual meaning.

Instead of the Next Sentence Prediction task, SOP is used to replace the NSP task in models such as ALBERT (Lan et al., 2020). Although originally designed for the improvement of understanding sentence relations, later research suggests that NSP may not enhance performance over downstream tasks as well as is needed. As it is more directly focused on ordering and coherence, SOP is the more relevant pretext task for training. Also, models trained with SOP as a pretext task perform better than those trained with more trivial tasks, indicating that the capacity to understand sentence order contributes positively to overall language understanding capabilities.

MosaiCLIP (Singh et al., 2023) employs a scene graph-based text decomposition strategy for the improvement of image-text matching. The authors come up with a coarse-to-fine contrastive learning framework to align images with multiple text sub-graphs in addressing compositionality challenges in vision-language tasks. This approach utilizes some hard-negative mining techniques, thus improving its performance significantly in many benchmarks and proving its capability to understand the hierarchical text relationship as well as the dynamics of the sentence order.

### 3.2.4. Auto-regressive Language Modeling (ARLMs)

ARLMs is a statistical approach to NLP for text generation. As the name "auto-regressive" suggests, it is based on predicting one word at a time, with previously generated words serving as context. Hence, the model relies on its past outputs to inform future predictions. For example, given the output of the model "I like," the next word in the output should be either "cats" or "chocolate" since probabilities have been learned on the training data. Several NLP tasks were proven to outperform traditional approaches with auto-regressive language models.

In question-answering and text-completion tasks, GPT-3 has already set benchmarks showing that ALM can indeed serve as a strong pretext for learning robust language representations. Auto-regressive models, like GPT-3, are widely recognized for achieving state-of-the-art performance across numerous NLP benchmarks; therefore, many discussions surrounding architecture and training methodologies have led to such a position.

, while MLM used in BERT focuses on masked word prediction for a sentence with both left and right contexts, ALM uses a strict left-to-right approach. This distinction makes ALM better suited to generative tasks where sequential coherence is crucial (Liao et al., 2020). It can predict words based on prior context, which helps make a subtle language structure and semantics, and can be used for several tasks other than text generation, like summarization and sentiment analysis

**Table 1.** What is Left Context and Right Context in the text?

| Sentence | Target Word | Left Context | Right Context |
|---|---|---|---|
| A cat sat behind the leaves. | Sitting | A cat | Behind the leaves |
| A honeybee approached a yellow flower. | approaching | A honeybee | A yellow flower |

### 3.2.5. Sentence Permutation

In Sentence Permutation, the task is to take a continuous span of text, thoroughly shuffling the sentences, and asking models to recover the original order. This task aims to improve understanding of sequential information and text coherence. Sentence permutation helps models better comprehend logical flow and narrative structure in texts. Training models on shuffled sentences helps them understand the links and dependencies that determine phrase order and coherence, preparing them for future NLP tasks.

Studies on text data augmentation suggest that using sentence permutation can enhance dataset diversity and classification accuracy. To improve model generalization across tasks, (Haralabopoulos et al., 2021) suggest applying additional permutations to sentence order in the training dataset.

Sentence permutation has also been used for paragraph generation. The PermGen framework (Yu et al., 2021) aims to optimize the predicted log-likelihood of paragraph outputs across all sentence orderings. This approach demonstrates how permuting phrase order improves content diversity in tasks like story and news production. Allowing models to produce different phrase structures during training improves the diversity and creativity of generated content.

# 4. Contrastive Learning approaches in Text-Image models

The majority of text-image models employ a variation of the InfoNCE loss to facilitate crossmodal alignment. The loss ensures that positive pairs (image-text pairs from the same sample) are closely aligned in the shared embedding space,, while negative pairs (non-matching imagetext combinations within a batch) are dispersed.

## 4.1. Cross-Modal Contrastive Learning

This category focuses on aligning representations from different modalities, such as textual and visual data, in a shared latent feature space. Cross-modal objectives form the backbone of many text-image models, enabling tasks like multimodal retrieval and zero-shot classification.

### 4.1.1. Symmetric Contrastive Objectives

Models like CLIP and ALIGN use symmetric contrastive objectives to align image and text embeddings in both directions. This means they optimize retrieval from images to text and vice versa, making them super effective for generalization. The symmetric loss formulations are robust against data variations and allow for scalable training on large datasets. CLIP works with carefully selected datasets of image-text pairs and does a great job generalizing to new tasks thanks to its strong cross-modal alignment. Its symmetric loss boosts alignment efficiency across both types of data. ALIGN takes this a step further by applying the symmetric contrastive loss to huge, noisy datasets, showing that sometimes, having more data can help tackle the issues that come with noisy label alignment.

The CLOOB framework (Fürst et al., 2022) further innovates these objectives by tackling the saturation problems often seen with InfoNCE. CLOOB introduces the InfoLOOB (InformationTheoretic Leave-One-Out Bound) objective, which not only aligns embedding, but also enhances uniformity in the shared embedding space. By improving covariance structures with modern Hopfield networks, CLOOB ensures that the distri, bution of embedding is wellbalanced, leading to better generalization for downstream tasks like fine-grained retrieval and multimodal reasoning.

### 4.1.2. Noise-Robust Learning

The use of noisy datasets, as in ALIGN, introduces a unique challenge for cross-modal learning. ALIGN demonstrates that even weakly labeled datasets (e.g., image alt-text pairs) can achieve robust alignment when trained at scale. By relying on large batch sizes and implicit negatives, it minimizes the impact of noise.

### 4.1.3. Fine-Grained Alignments

Some models extend contrastive learning to improve the detail of interactions between image and text modalities by aligning smaller important aspects, like image patches or text objects. Rather than considering the entire image or text as one entity, they align smaller, significant elements, such as particular image segments or textual items. This method goes beyond just global embedding, allowing models to capture finer details and relationships that enhance retrieval accuracy and reasoning skills. For example, instead of connecting a whole image to a sentence, these methods could link particular sections of an image (such as a dog or a cat) to the relevant words or expressions in the text. This comprehensive alignment is particularly beneficial for tasks demanding a profound comprehension, like object detection or regionspecific image captioning, where recognizing localized relationships is essential.

, while these methods are effective, they entail a significant computational cost. Aligning at the level of patches or objects requires additional computations, such as attention mechanisms, to accurately compare localized features. Additionally, handling and processing the increased number of embeddings for different patches or objects within a single image takes significant memory and computational resources.

Despite these challenges, fine-grained alignments are quite useful for boosting retrieval accuracy and enhancing task-specific performance. They capture fine details, enabling models to tackle complex queries that require detailed comprehension (like "find an image with a dog under a tree") and produce outputs that are more contextually relevant, making them a promising path for advanced multimodal applications.

## 4.2. Unimodal Contrastive Learning in Text-Image Models

On the other hand, although cross-modal objectives aim to synchronize various modalities, unimodal Contrastive learning emphasizes preserving the distinct characteristics of each modality in text-image models. This method ensures that the model retains comprehensive representations of images and text separately.

### 4.2.1. Self-Supervised Learning (SSL)

In unimodal contrastive learning, self-supervised objectives, like those in SLIP, aim to align different augmented views of the same input image. For image learning, various augmentations (such as cropping, flipping, or jittering) of the image are created and are treated as positive pairs,, while all other instances in the batch are considered negatives. This process helps the model learn the visual representations and improves accuracy in downstream tasks like classification and segmentation. SLIP utilized self-supervised learning for images with cross-modal alignment, ensuring that both image-specific (unimodal) and multimodal tasks can utilize the learned features.

### 4.2.2. Balancing Modality-Specific Features

In multimodal systems, it is crucial to avoid feature collapse, which occurs when the model becomes overly focused on cross-modal alignments and begins to neglect the distinct characteristics of each modality. The unimodal contrastive learning method in COOKIE (Wen et al., 2022) addresses this problem by applying different losses for every modality. For example, the Visual Contrastive Loss (VCL) preserves complex visual characteristics for tasks like object detection and image retrieval, whereas the Textual Contrastive Loss (TCL) ensures that language representations are deep and meaningful for sentiment analysis and natural language understanding.

CLOOB employs modern Hopfield networks to ensure that these distinctive features for each modality are preserved even, while the embeddings are synchronized in a shared space. Each domain's accuracy is maintained by properly retaining high-dimensional representations.

## 4.3 Key Trends in Contrastive Learning Approaches

### 4.3.1. Scaling with Noise

The scalability of contrastive learning is one of its standout features, and being able to manage noisy data is crucial when dealing with large, weakly labeled datasets. Take ALIGN (Jia et al., 2021) as a prime example; it shows how working at a large scale can help mitigate the noise often found in datasets like image alt-text pairs pulled from the internet. ALIGN proves that even without extensive cleaning or curation, using over a billion noisy image-text pairs still allows the model to learn representations. This success comes from the sheer volume of data, which helps the model spot consistent patterns and effectively disregard noisy or mislabeled samples by averaging across large batches.

On the other hand, CLIP (Radford et al., 2021) uses a high-quality dataset. This strategy tackles the issues due to noise in data,, but this results from the dataset being smaller. The careful selection ensures that each image-text pairing is semantically relevant, leading to more accurate representations, even with limited data., while this method reduces the need for vast amounts of data, it also creates dependency on manual annotation, which can be time-consuming and hard to scale.

CLOOB (Fürst et al., 2022) takes a different approach by concentrating on enhancing the consistency of embedding rather than just focusing on data volume. It uses InfoLOOB, which tackles the challenge of embedding saturation, ensuring that representations stay evenly spread out in high-dimensional space, even when trained on smaller or noisier datasets. This approach improves the accuracy of image-text contrastive learning models to some extent.

### 4.3.2. Batch Dynamics and Negative Sampling

Negative sampling plays a crucial role in contrastive learning because it helps distinguish between positive and negative pairs. CLIP and ALIGN use in-batch negatives, where all non-matching image-text pairs in a batch are counted as negatives. This approach effectively takes advantage of the variety of negatives without needing extra computation outside the batch. However, as the batch size increases, the computational demands of contrastive learning can skyrocket due to the quadratic complexity involved in calculating similarities for all pairs in the batch. But larger batch dramatically increase computation due to quadratic complexity in pair similarity calculations. This situation can create a bottleneck in terms of both memory usage and computational load.

One major challenge in contrastive learning is when hardware limitations cap the batch size, which can limit the diversity of negative samples. Small batches limit comparisons, risking less effective representations as models might overfit to few negatives. On top of that, the issue of hard negatives adds another layer of complexity to training. Hard negatives are pairs in the batch that are categorized as negative pairs, but are semantically closer to the positive pair (for instance, images of similar scenes, but with different captions). It is important to categorize these samples as positive pairs to encourage models to learn accurate general representations.

However, finding and dynamically including hard negatives during training can be quite resource-intensive, often requiring extra resources or specialized sampling techniques.

### 4.3.3. Temperature Scaling

Temperature scaling is essential in contrastive learning as it determines the intensity of the similarity distri, bution in the loss function. It does so by adjusting similarity scores before applying the softmax function. The temperature parameter (τ) is crucial: when configured to lower values (τ<1), it produces sharper distri, butions that enhance the contrast between positive and negative pairs. On the other side, higher values (τ>1) smooth out the contrast between pairs, leading to more generalized representations. In models that deal with text and images, temperature scaling helps balance out the differences in modality, like the tendency for text embedding to cluster more densely than image embedding, which is crucial for maintaining effective learning dynamics. Furthermore, getting τ just right boosts training stability by stopping extreme similarity scores from taking over the loss calculations and helps reduce the risk of overfitting on those tricky hard negatives. Some methods even treat τ as a learnable parameter, allowing the model to adjust dynamically based on the dataset's characteristics and the training progress, which ultimately enhances alignment and generalization.

# 5. An Architectural overview of Contrastive Learning based Text-Image models

## 5.1. Dual Stream Architecture

Dual stream architecture features distinct routes for handling both images and text,, but these streams are linked to help align the different modalities. COOKIE (Wen et al., 2022) utilizes a text-aligned visual transformer encoder that efficiently aligns the distri, butions of visual and textual attributes, improving cross-modal education. Although textual characteristics from transformers similar to BERT (Devlin et al., 2019) are generally comprehensive, visual attributes derived from Convolutional Neural Networks (CNNs) such as ResNet (He et al., 2015) tend to be more localized and frequently miss that global viewpoint. The Text-Aligned Visual Transformer Encoder addresses this gap by utilizing a transformer encoder applied to the visual features, facilitating global attention mechanisms that better capture the relationships across visual elements. In contrast to separate processing paths, the Weight-Sharing Transformer Encoder employs a common set of transformer weights utilized for both visual and textual features. This weight-sharing mechanism ensures that the encoder develops an accurate representation, prompting image and text embedding to concentrate on similar contextual details. By leveraging shared parameters, Weight-Sharing Transformer Encoder facilitates efficient knowledge transfer between modalities, leading to improved semantic coherence.

### 5.2. Dual Encoder Architecture

Most multimodal text-vision architectures process images and texts separately using dedicated encoders like CNNs or vision transformers for images and Transformers for texts. CLIP, SLIP (Mu et al., 2021), and ALIGN use ResNet with global pooling or the Vision Transformer (ViT) (Dosovitskiy et al., 2021) as their image encoders and employ a Transformer-based text encoder. Both modalities are processed in different encoders; however, it does not use any architectural connection to overlay the modalities, instead using a shared latent space where both image and text encodings from encoders are mapped.

The shared latent space is a common embedding space where representations from two different modalities, such as images and text, are aligned. The concept is foundational to their capabilities in handling cross-modal tasks effectively, like image-text retrieval, zero-shot classification, and multimodal reasoning. Both modalities are processed by their respective encoders and projected into this high-dimensional space, where Image-text pairs with similar semantics are placed close together,, while dissimilar pairs are farther apart. The alignment is achieved using a contrastive learning objective that maximizes similarity for positive pairs and minimizes it for negatives. This space enables tasks like image-text retrieval and zero-shot classification by allowing direct comparisons using cosine similarity. Its design ensures scalability, efficiency, and semantic coherence, making it foundational for multimodal learning.

### 5.3. Single Tier Architecture

Single tier refers to a single Transformer-based architecture to process and encode both image and text data. Both images and text are fed into the same pipeline and processed together in shared layers without explicitly separating them into independent processing tiers. After dividing images into patch embedding and embedding tokenized text sequences into token vectors, LIMoE (Mustafa et al., 2022) incorporates Mixture-of-Experts (MoE) layers within the shared Transformer backbone to provide sparse activation. MoE is a machine learning approach that partitions AI models into specialized sub-networks (experts),, while each focuses on a certain subset of the input data to perform a combined task to collaboratively enhance tasks (Z. Chen et al., 2022). LIMoE uses a sparsely activated MoE architecture, meaning rather than using the entire network for every input, only a small subset of the model's experts is activated for any given input. This means that in LIMOE, when the input is an image, only certain experts will be activated,, while for text, a different subset of experts will be activated.

**Table 2.** Architectural Comparison

| Features | Dual Stream Architecture | Dual Encoder Architecture | Single Tier Architecture |
|---|---|---|---|
| **Scalability and Efficiency** | Efficient knowledge transfer between modalities via shared parameters; improved semantic coherence. | Highly scalable due to independent encoders and shared latent space. | Sparse activation causes a reduction in computational overhead by activating only relevant experts for specific inputs. |
| **Use Cases** | Cross-modal learning tasks that require semantic coherence, for example, multimodal reasoning. | Image-text retrieval, zero-shot classification, and other tasks leveraging shared latent space alignment. | Suitable for complex reasoning tasks with sparse expert activation mechanisms. |
| **Examples** | COOKIE (Wen et al., 2022) | CLIP, SLIP (Mu et al., 2021), ALIGN using ResNet or Vision Transformer (ViT) | LIMoE (Mustafa et al., 2022) |

# 6. Integrating Self-Supervised Contrastive Learning into Supervised Learning Tasks

## 6.1. Data labeling in Supervised Learning

Supervised learning is the primary approach used in machine learning. The models are trained on labeled datasets; that is, input-output pairs to predict the outcome of certain features in the input. Every input data point comes with a corresponding label (output), such that the model can be thought to learn their mapping. Supervised learning tasks include classification and regression.

High-quality labeling datasets pose huge challenges and significantly influence machine learning models' performance and reliability. Usually, there is a need to have domain-specific knowledge in labeling tasks since Labor-Intensive processes require quality assurance and raise ethical concerns. Images labeled by humans can be inconsistent or subjective, leading to ambiguity in annotations and eventually a low-accuracy model. Other than this, as datasets grow in size, scaling the labeling process becomes increasingly challenging.
This brings forth the need for the integration of Self-Supervised Learning into Supervised Learning tasks.

## 6.2. Methodologies for Integration

### 6.2.1. Zero-Shot Learning (ZSL)

Zero-shot learning enables models to classify concepts without them having to exist in the training process. Rather than restricting itself to labeled examples for each class, the ZSL paradigm uses semantic information such as attributes or textual descriptions to build relationships between classes the model was trained on and unseen classes it encounters at inference time. It would, therefore, be possible to generalize the model's knowledge towards new categories without requiring more labeled data for these categories.

As a result, combining zero-shot learning with supervision provides models with the opportunity to utilize labeled and unlabeled data. This will, therefore, enhance the model's ability to generalize tasks by maximizing the usage of labeled data or when labeling is not straightforward (Pourpanah et al., 2022).

#### 6.2.1.1. Transfer Learning Techniques

Transfer learning bridges the gap between ZSL and the corresponding supervised tasks. It reveals that by knowing the features being shared, knowledge would be transferred to unknown classes for classification with well-trained models, therefore bringing ZSL about without extensive retraining. Integration of Zero-shot Learning with Supervised Learning tasks via transfer learning involves expanding label coverage and improving generalization (Pourpanah et al., 2022).

#### 6.2.1.2. Attribute Selection

Research shows that combining ZSL (Yuchen Guo, n.d.) and attribute-based classification techniques models can infer unseen classes. The model achieves it through the utilization of semantic attributes related to observed classes. The finer set of attributes increases the ability to classify over different datasets since all attributes do not contribute equally to ZSL performance.

#### 6.2.1.3. Hierarchical Classification

In ZSL, semantic representations describe both seen and unseen classes, whereas Hierarchies enhance embedding by considering parent-child relationships, improving the model's understanding of label proximity. Hierarchical label sets provide a structured way to address the challenges of Zero-Shot Learning, particularly in handling fine-grained labels. The CHiLS (Novack et al., 2023) method uses these hierarchical relationships to predict high-level categories first and then refines the predictions to more specific labels within the hierarchy, improving zero-shot classification performance.

### 6.2. Linear Classification

Linear classification uses a straight line or hyperplane to classify data points, which splits the input space into two or more regions, and the adopted model will attempt to find the linear decision boundary that might further split input features into classes. The approach is to label data according to which side of the line or hyperplane it falls on. The simplest form of linear classification tries to find the best function to separate classes. For example, for 2D data, a line; for 3D data, a plane; and so on.

This role is extremely important for evaluating and using the learned features based on selfsupervised learning, especially in models that are designed with contrastive text-image models. Following this contrastive learning procedure, a linear layer is often added to the frozen encoder outputs to adapt the model to the downstream task, including feature extraction and zero-shot learning. Adding linear classification to the contrastive learning setting improves model performance when the training process is carried out on unlabeled data.

### 6.2.1. Contrastive Learning

The contrastive learning is exploited by the SimCLR (T. Chen et al., 2020) and MoCo (He et al., 2020) frameworks to learn strong feature representations that work well in follow-up supervised learning tasks.

SimCLR is a self-supervised learning method that concentrates on maximizing the consistency between augmented views of the same image and reducing the correspondence between various image pairs, without depending on positive and negative label images. It applies the NTXent normalized temperature-scaled cross-entropy loss, which prevents noise and efficiently separates pairs within the learned representation space. These SimCLR-learned representations can be employed directly in linear classifiers for downstream tasks. It minimizes the model's reliance on large, labeled datasets by optimizing the representations through SSL, thereby enabling efficient training even with very limited labeled data. The ability to fine-tune from a middle layer of the projection head further improves performance, yielding improvements in both linear evaluation and fine-tuning scenarios.

MoCo v3 (X. Chen et al., 2021) is another self-supervised learning framework that further extends its predecessors by introducing several enhancement techniques to stabilize the training of vision transformers., This, too,, while improving the quality of learned representations. It is unimodular and employs a dynamic dictionary of negative instances, enabling more effective contrastive learning by strengthening agreement between positive augmented-pairs views of the same image, while reducing agreement with negative views from different images.

Models learned using MoCo v3 representations have exhibited higher accuracy in linear classification compared to those that learn from other self-supervised approaches. Indeed, empirical work shows that MoCo v3 produces significant performance gains over many architectures, including Vision Transformers (ViTs) (Khan et al., 2023). For instance, the linear probing accuracy for MoCo v3 with ViT-B achieved 76.7%,, while larger models like ViT-L and ViTH gained 77.6% and 78.1%, respectively.

### 6.2.2. Linear Probing (Evaluation Method)

Models such as CoCa (Tosh et al., 2021) and CLIP work with linear probing. In linear probing, the linear classifier is trained over the representations derived with self-supervised approaches. It becomes an excellent technique to evaluate the quality of the features learned and utilize them for more downstream tasks.

(Huang et al., 2024) considers the advancement of linear probing for the particular few-shot adaptation of CLIP. The authors propose a general linear probing formulation that includes learnable text embedding functions in the classifier weights for enhanced performance in fewshot learning. The research indicates how linear probing can be optimized and refined for optimal performance when used with models such as CLIP.

Just like CLIP, CoCa also employs linear probing to probe its learned representations. For this purpose, a linear layer is added on top of the backbone of the model after its contrastive training stage. The training process is continued for the layer, while keeping the backbone frozen. This lets it adapt to specific downstream tasks without affecting the underlying representation that was learned during pre-training.

Linear probing is widely employed as a benchmark over self-supervised models, for the performance of the linear classifier reflects the effectiveness and generality of the pre-trained features. It becomes particularly useful when labeled data is limited, tapping into the rich, transferable representations learned during self-supervised training for strong performances across a variety of tasks.

### 6.3. End-to-end Fine Tuning:

End-to-end fine-tuning is a very powerful technique to incorporate self-supervised learning representations, particularly from models via image-text contrastive learning, into the task at hand. This method adapts all of the pre-trained models for the specific task so that it can refine its learned representations for best performance.

After pre-training, a linear classifier can be trained on top of the representations learned by SimCLR. That is, fine-tune the classifier from labeled data with the advantages of SSL to learn specific tasks. The paper indicates that with representations from SimCLR, a linear classifier trained on top performs competitively with fully supervised models. It will demonstrate how SSL can significantly decrease dependency on labeled datasets (T. Chen et al., 2020).
To optimize for supervised tasks, after pre-training, CLIP can be optimized for a particular supervised task by placing a linear classifier on top of its acquired representations. This allows knowledge learned from SSL to be transferred effectively into a variety of downstream applications. CLIP's architecture enables it to work effectively on the most varied tasks with little additional training, opening the door to combine SSL with fully supervised learning pipelines (Radford et al., 2021).

## 7. Challenges with Integration and the Solutions

Combining SSL into supervised learning problems is advantageous,, but is beset by several challenges, which include Data Quality and Diversity, Computational Complexity, Generalization, and Adaptability.

### 7.1. Diversity and Quality of Data:

The performance of SSL methods has a great dependency on the quality and diversity of the unlabeled training dataset. The learned representations may not be accurate or generalizable if the training dataset was biased or noisy. In any imaging application, low-quality annotations may cause model prediction biases to skew significantly, making it less accurate. Furthermore, large class imbalances in a dataset also negatively impact performance when minority classes are largely underrepresented. This may be especially challenging in certain domains such as medical imaging (Rauf, Khan, & Khan, 2024), because of the rarity of a few diseases.

**Solution:** Data Diversity through data augmentation and better data acquisition strategies. Data augmentation, such as cropping, rotation, color adjustment, and other transforms, can create different views of the same dataset. It helps train the model on how features are invariant to input changes. Then, devise strategies for acquiring data from different sources and environments and ensure the data is balanced for all classes.

### 7.2. Computational complexity

The SSL methods, especially the ones with advanced architectures such as transformers, increase computational complexity. Self-attention in transformers involves several steps that are additive in nature and hence contribute to the overall computational complexity. This can be a major challenge, especially when dealing with large datasets, since training models on such datasets is computationally intensive, especially when using complex architectures.

**Solution:** Networks can be optimized for performance, and better strategies can be employed to optimize their efficiency. For example, distri, buted training, along with a more efficient architecture model, can efficiently address computational problems.

Refining contrastive learning methods also decreases computational complexity. Techniques like SimCLR and SwAV (Caron et al., 2021) use efficient loss functions for contrastive loss that are less computationally intensive than standard losses in contrastive learning (Khan et al., 2024).

### 7.3. Generalization and Adaptability:

If SSL models are not designed to adapt, then they fail to generalize to new tasks or environments; hence, their applicability turns out to be very limited to real-world scenarios where the change in objectives of the tasks occurs frequently, and changeability is a necessity for their network scenarios.

**Solution:** Solutions to this challenge have been extensively discussed in (Yang et al., 2024).
Continual learning or lifelong learning is training a model on a sequence of tasks without forgetting the previously learned knowledge. Therefore, combining SSL with continual learning would allow the model to be adapted continuously to new tasks and environments.

Hybrid Approaches, i.e., combining SSL with other machine learning approaches like meta learning, few-shot learning, reinforced learning, or online learning, can add greater adaptability to the model. For instance, the representation of a model may be improved by training on SelfSupervised tasks like colorization and depth prediction and then fine-tuned for a supervised task like semantic segmentation. The runtime of the supervised task does not increase,, but it is improved.

Another solution is to train a model in multiple tasks at the same time. This improves the model's adaptability as well as its generalization capabilities. By combining SSL with multi-task learning, shared representations that are applicable across diverse tasks can be learned by this model.

# 8. Applications of Contrastive Learning based Text-Image Models

## 8.1. Image Classification

Image classification identifies and categorizes objects in an image. This can be achieved with the help of models trained with labeled data, where every image is associated with a class label. Most traditional techniques for image classification tend to fail when they encounter classes that are outside their training.

CHiLS (Novack et al., 2023) addresses limitations in zero-shot classification, which classifies images without target class labels.

CHiLS uses hierarchical structures for class labels and enables the sub-classification of classes with semantic relations among them. It opts for flat class labels over subclasses. Subclass can be generated by exploiting a given hierarchical label structure or directly by a language model like GPT-3 to be able to automatically generate meaningful subclasses. The model then uses the standard zero-shot classification process with these subclasses as labels, thus treating them as primary targets to predict. Finally, the predicted subclass is mapped to its parent class to give the final classification result.

The hierarchical approach enables CHiLS to boost accuracy in many datasets, especially those with implicit semantic hierarchies, since it provides richer and more informative definitions of classes.

SimCLR has also received much attention for image classification tasks, especially under weakly labeled data availability. It relies on strong techniques in data augmentation and incorporates a learnable nonlinear transformation between learned representations and contrastive loss.

Having more negative samples helps the model to learn better distinctions between classes. Therefore, SimCLR uses larger batch sizes and more training steps compared to traditional supervised learning methods.

It used to evaluate SimCLR with multiple imaging modalities which were based on Retina Fundus, Blood Cells, Colon Pathology, and Dermatology in the medical field for examining the potential ability for enhancement in classification tasks for improved images of medical cases especially with fewer labeled data sets and sometimes unbalanced classes,, but they obtained impressive increments as high as 30.6% accuracy above baseline supervisory metrics. This identifies the potential of SimCLR in domains where labeled data is scarce, but large amounts of unlabeled data are available.

The key contribution was the development of an augmentation sequence that showed consistent improvement in performance over the original SimCLR framework. This goes to show how tailored data augmentation strategies are important to the improvement of model robustness and accuracy (Azizi et al., 2021).

BYOL (Grill et al., 2020) is a self-supervised learning technique designed to learn visual representations without relying on negative samples (dissimilar images/objects). It utilizes two neural networks called online and target networks. Online Network: It consists of an encoder, a projector, and a predictor. It processes augmented versions of an image and predicts its representation from the target network, while target networks provide a stable representation, online networks align with it by facilitating effective self-supervised learning. The architecture of the target network as as the same as the online network,, but it maintains different sets of weights, updated parameters using a slowmoving average of the online network's weights.

The concept of BYOL is based on a teacher-student learner mechanism where both networks train to match the similarity of augmented images. The technique does not use the dissimilar pairs of images/objects (negative pairs) like SimCLR and MoCo,, but instead focuses on maximizing similarity between augmented images of the same class.

BYOL achieved a top-1 classification accuracy of about 74.3% on the ImageNet dataset by using the ResNet-50 architecture. It further improved its performance to 79.6% by using a larger ResNet model (a deep network with too many layers). These results demonstrate the effectiveness of the BYOL technique in learning visual representations.

Studies have demonstrated that BYOL might improve performance in some applications, for example, mosquito classification. The main objective of this work is to improve mosquito classification accuracy. This is crucial in most public health contexts where proper identification of a mosquito species may contribute to better control over diseases spread by these insects.

Self-supervised techniques like BYOL could aid public health by monitoring mosquito populations and preventing malaria and dengue (Akter et al., 2021).

## 8.2. Visual Q&A

Visual question answering is an interdisciplinary task integrating the fields of computer vision and natural language processing, to enable machines to understand a question and answer it given an image. The processing capacity of the CLIP model makes it useful in any VQA application by integrating the image and text representations. Visual Mind (Sonawale & Shaikh, n.d.) outlines a holistic approach for VQA by using the CLIP model, particularly extending its capabilities with more layers and exploiting the VizWiz dataset designed for people with visual impairments. Results the improved VQA-CLIP model is comparable with other state-of-the-art techniques: Accuracy metrics of the different architectures (ResNet-50, ResNet-101, Vision Transformer) are improved dramatically on VQA tasks. High answerability accuracy across models means that they correctly filter the unanswerable ones.

PubMedCLIP (Eslami et al., 2023) is a fine-tuned adaptation of the original CLIP, especially for medical-related VQA tasks with the aid of data coming from articles on PubMed. The improvement process was to improve its performance in VQA for medicine-related tasks. The authors ran experiments on two benchmark datasets for MedVQA to measure the performance of

PubMedCLIP compared to existing models. The authors compared PubMedCLIP to state-of-theart approaches, including model-agnostic meta-learning networks, which are pre-trained exclusively on visual data. Their results showed that PubMedCLIP performed better, gaining up to 3% better accuracy in general performance over MAML networks.

Research on Unbiased VQA using Contrastive Learning, or CL-VQA, has gained traction as a method to address the inherent biases present in traditional VQA models. In Robust Visual Question Answering by Polarity Enhancement and Contrast research (Peng & Li, 2024), which proposes highlighting the contrast between correct answers and positive and negative predictions. The model comprises Answer Visual Attention Modules that make positive predictions through processing the input question and image, and a Dual Channels Joint Module that makes negative predictions through using linguistic prior knowledge. The method comprises a newly formulated loss function with both positive and negative predictions added together with the ground truth, achieving a performance grade of 61.24% on the VQA-CP v2 dataset and improving performance on the VQA v2 dataset. This demonstrates the model's effectiveness in mitigating language bias without sacrificing overall performance.

## 8.3. Content Moderation

Content moderation in image-text contrastive learning models is an emerging area of research, focusing on the challenges and methodologies for ensuring that generated content adheres to specific ethical and community standards. This involves using advanced models such as CLIP and others to manage and filter out inappropriate or harmful content effectively.

Although state-of-the-art models achieve good performances for commonplace concepts, they fail to handle fine-grained concepts that are underrepresented in their respective datasets, which are used for training purposes. Based on this limitation, newer efforts have been proposed involving the retrieval-based approach through better alignment of image-text representation by incorporating external memory support for fine-grained retrieval of knowledge.

It uses a lightweight fusion transformer that sits atop a frozen CLIP model for improving the performance across a diverse set of tasks without requiring too much additional pre-training. The information retrieval from external memory by this fusion transformer gets integrated into the model's embedding at the time of inference, which enables the system to process complex and nuanced queries.

The paper also explains the performance of unimodal and cross-modal fusion techniques. The performance of the cross-modal fusion can be stated to be larger than that of the unimodal equivalent in leveraging complementary information in the textual and visual modalities to align and better comprehend multimodal data (Iscen et al., 2024).

Even with advancements, some challenges do emerge in moderating image-text model outputs for issues such as the Complexity of User Prompts. Objectionable content can be created by users with innovative or roundabout prompts that could bypass straightforward keyword filtering. Robust moderation needs to take such complexity in the form of sophisticated synonym and context-based policies into consideration. Hence, Model Robustness becomes important to keep models generalizing reasonably well across numerous settings with high performance in detecting objectionable content. This is a matter of constant updating and training on many datasets to encompass changing language and visual form (P. Wang et al., 2024).

## 8.4. Image and Video Retrieval

The retrieval-augmented approach has been offered in RA-CLIP (Xie et al., 2023). This model can leverage a reference set of image-text pairs to add richness to the representation of input images and thus serve as a "cheat sheet" in the case of recognition tasks. The author of the paper describes this by using the analogy of an "open-book exam". The experiments demonstrate that RA-CLIP significantly outperforms the vanilla CLIP model across multiple benchmarks:

| | |
|---|---|
| Zero-Shot Image Classification | +12.7% improved performance |
| Linear Probe Image Classification | +6.9% improved performance |
| Zero-Shot ROI Classification | +2.8% improved performance |

Compared with CLIP, the key innovation in RA-CLIP is of Retrieval Augmentation Module (RAM). The process involves the use of cosine similarity to find the closest matching embedding, which is referred to as selecting the top-K most similar image-text pairs. Finally, the model improves the result via fusion of Image Augmented Embedding, Test Augmented Embedding, and CLIP's Image embedding. This is represented in equation (1).

$$v_i' = v_i + a_i^T + a_i^I \quad (1)$$

Where,

$v_i$ : CLIP's Image embedding,

$a_i^T$ : *Text* Augmented embedding from RAM,

$a_i^I$ : Image Augmented embedding from RAM.

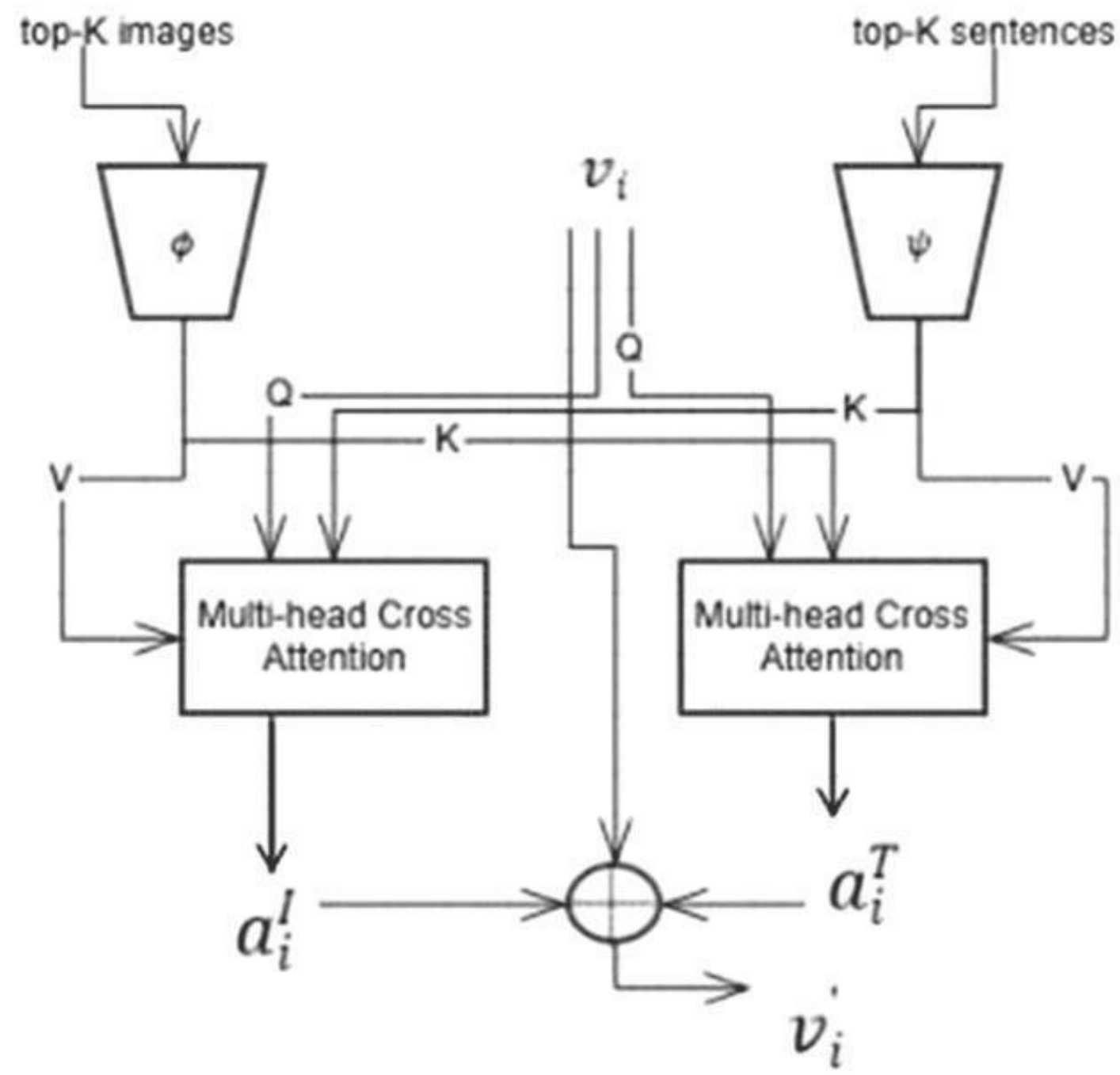

**Figure 9.** Retrieval Augmentation Module (RAM)

The RA-CLIP framework introduces a method with which relevant image-text pairs are retrieved from a hold-out reference set to complement the representation of input images. This process allows more context to be used within the model during inference, which acts as a retrievalbased system for improving image understanding.

Contrastive learning approaches such as SimCLR and MOCO reveal their capacity to lower intraclass variability, while adequately retaining instance-level relations relevant to retrieval tasks. Empirical comparisons based on benchmark sets such as Oxford5k (Philbin et al., 2007) and Paris6k (Philbin et al., 2008) have shown SimCLR to be better in achieving high mean Average Precision values. The implications are that contrastive learning methods have been found to improve instance discrimination and often compete with or even outperform the performance of supervised baseline models in instance-based retrieval tasks (Krishna et al., 2021).

VideoCoCa (Yan et al., 2023) is a video-text model explicitly created for video-text tasks, utilizing a pre-trained image-text contrastive captioner (CoCa) to facilitate effective zero-shot transfer. It includes generative and contrastive attentional pooling layers from CoCa on top of flattened frame embedding for the best performance up to date in zero-shot video classification and text-to-video retrieval. Lightweight fine-tuning helps further boost its application for tasks such as video question answering and captioning that obtain robust results with minimal need for overfitting. VideoCoCa thereby reuses existing trained models to efficiently and effectively carry out multimodal understanding.

VideoCLIP (Xu et al., 2021) is a state-of-the-art video-text understanding framework with new contrastive learning methods. It drops the need for strictly aligned video-text pairs. It can work on loosely overlapping clips, facilitating much more flexible and effective multimodal associations, thereby boosting the number and diversity of positive pairings seen during training. Still, challenging examples from other videos further drive the model's discriminative capabilities between video and text representations.

Experiments show that VideoCLIP performs strongly in zero-shot settings, and in some cases, also outperforms fully supervised methods. In tasks such as text-video retrieval, it dominates all existing zero-shot methods and even surpasses models that require large amounts of labeled data for fine-tuning.

## 8.5. 3D Vision

MixCon3D (Gao et al., 2024) aims to improve the 3D representation learned using contrastive learning techniques. It aims to develop 3D object-level representations with the help of multiview rendered images and point cloud data. The framework uses joint representation alignment techniques, which are the concatenation of features and contrastive learning.

In Concatenation of Features, point clouds are concatenated to form a joint representation, and features are extracted from multi-view images. This is an important process as it makes crossmodal alignment robust in the sense that the 3D object representation maps to its textual description correctly. Contrastive learning is employed to align these joint representations with text embedding. It works by decreasing the similarity of the representations of various objects and increasing the dissimilarity for other objects, thus making it enhance the ability of the model to learn and produce 3D object descriptions.

MixCon3D has made notable gains over state-of-the-art techniques on several benchmarks, such as a 5.7% gain on the Objaverse-LVIS dataset (Gupta et al., 2019), as well as improved performance on the ScanObjectNN dataset (Uy et al., 2019). It has additionally been tested on many other tasks, such as Text-to-3D Retrieval and Point Cloud Captioning.

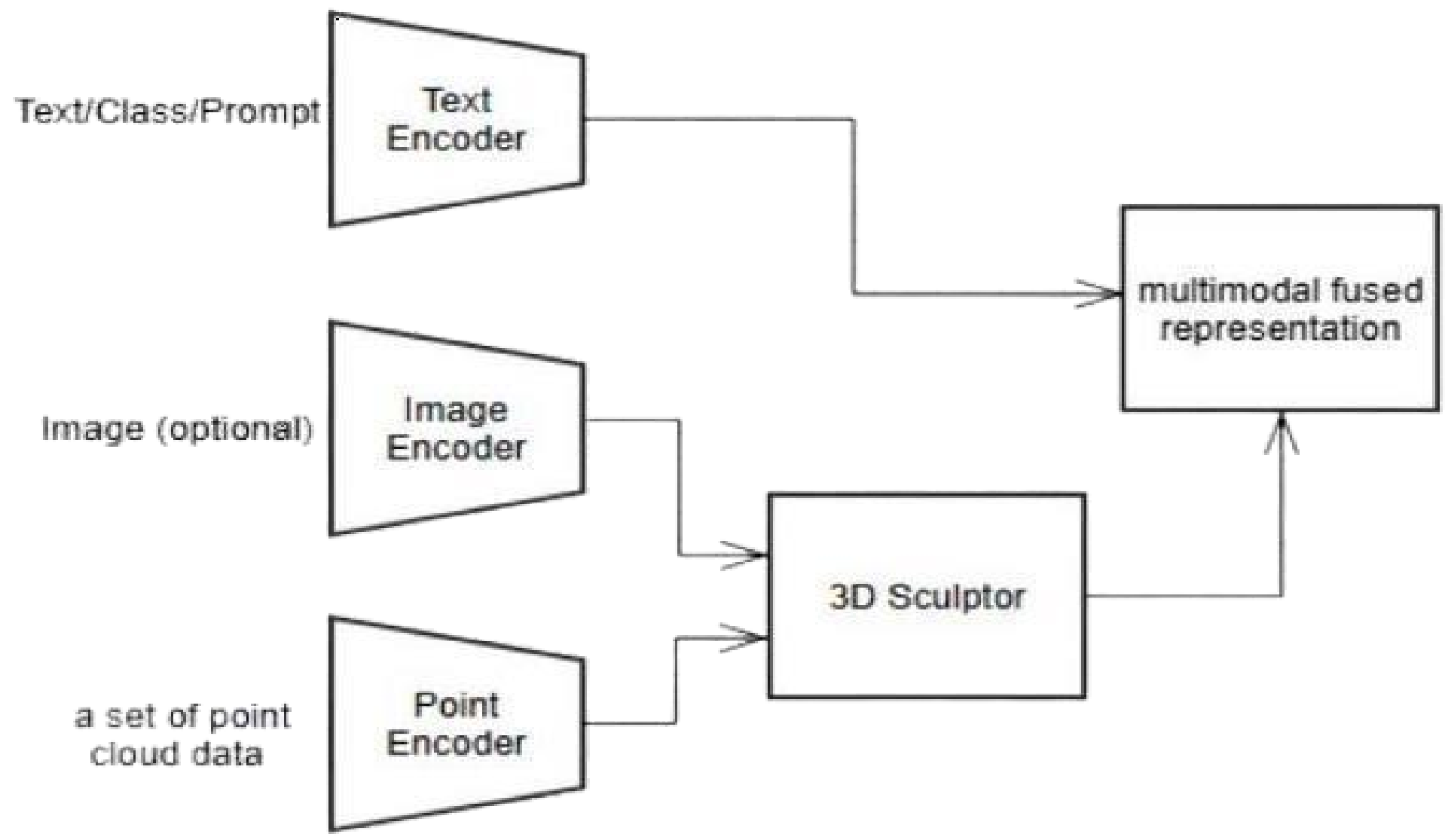

**Figure 10.** Summary of MixCon3D Framework

## 8.6. Medical Imaging:

One interesting use of Self-supervised multimodal contrastive learning in medical imaging has been discussed in "Benchmarking Vision-Language Contrastive Methods for Medical Representation Learning". The paper mentions 2.8 million image-text pairs across multiple datasets and evaluation on 25 downstream tasks. Images included data such as X-rays and MRIs,, while text included radiology reports. It uses contrastive learning to align textual descriptions with visual features, enabling tasks like zero-shot classification, linear probing, retrieval, visual question-answering, and reasoning.

**Table 3:** Summary of explored Contrastive methods for medical representation learning.

| **Explored Methods** | **Description of Explored Methods** | **Findings** |
|---|---|---|
| **Transferability Analysis** | Freezing parts of pre-trained generaldomain image and text encoders to assess how well these representations adapt to medical tasks. | General-domain representations can be transferred to the medical domain, especially when the image encoder is partially frozen. |
| **Unimodal Learning** | Combining multimodal contrastive loss with unimodal contrastive learning techniques like SimCLR (image augmentations) and masked image patches. | Adding unimodal representation learning does not improve multimodal representation learning in medical downstream tasks. |
| **Feature Granularity** | Using CoCa (Contrastive Captioners) for fine-grained feature learning through simultaneous image-text contrastive learning and image captioning. | Fine-grained representation learning enhances multimodal medical representations. |

# 9. Challenges and Future Direction

Popular frameworks like SimCLR, MoCo, and BYOL are still influential, but now new research builds on them by improving supervisory signals, leveraging attention mechanisms, and optimizing contrastive loss functions for better downstream performance and reduced need for labeled data. "SynCo: Synthetic Hard Negatives for Contrastive Visual Representation Learning", builds on the MoCo framework and addresses the challenge of hard negatives. It discusses generating synthetic hard negatives on the fly using six innovative strategies, including interpolated, extrapolated, mixup, noise-injected, perturbed, and adversarial negatives. Creating Synthetic Hard Negatives means making artificial hard negative samples. They help models to learn more precise and discriminative features.The question that arises here is, are creating Synthetic Negatives related to Data Augmentation? Data augmentation creates different positive views of the same original data point (using cropping or color-changing an image, etc.) while synthetic negatives generate hard "different" versions that are close but not the same as the original to improve contrastive learning.SynCo enhances downstream performance on ImageNet and transfer tasks by augmenting the contrastive learning process with these synthetic negatives.It generalizes beyond MoCo and claims applicability to SimCLR, BYOL, and other contrastive methods, helping achieve better robustness and efficiency while reducing reliance on labeled data. The approach also achieves faster training and better feature representation distribution.(Giakoumoglou & Stathaki, 2024)

The challenge currently being discussed in the self-supervised contrastive learning (CL) studies includes the gap between self-supervised contrastive loss and supervised variants that still exists. The gap tends to increase slightly during training but remains bounded by the theoretical limit. For datasets with fewer classes or uneven class size distributions, the gap can be more pronounced because the probability of sampling same-class negatives in the DCL denominator increases. Therefore creating SSL approaches that maintain efficacy under class imbalance and more complex data conditions and designing loss functions and training regimes that minimize the gap between supervised and self-supervised contrastive objectives more rapidly and consistently is the directly being taken. (Luthra, Yang, & Galanti, 2025)

## 10. Conclusion

In this survey paper, we provide a comprehensive review of self-supervised contrastive learning for multimodal text-image analysis. We clarified essential terminologies, provided a structured categorization of recent approaches. By examining evolving architectures and methodologies, we describe the effective contrastive learning frameworks such as CLIP, ALIGN, and CLOOB. We also discuss how Self-supervised contrastive learning has become a foundation for numerous downstream applications, such as cross-model retrieval, image captioning, and visual question answering. With that, we were able to address the practical impact and versatility of self-supervised contrastive learning in text-image multimodal tasks. At the end, we conclude the paper with trends like hybrid architectures and adaptive temperature scaling emerging in this field.

Our survey provides a foundational understanding and a coherent reference to guide future research and development in this area of study.

## References

Akter, M., Hossain, M. S., Ahmed, T. U., & Andersson, K. (2021). Mosquito Classification Using Convolutional Neural Network with Data Augmentation. In P. Vasant, I. Zelinka, & G.-W. Weber (Eds.), Intelligent Computing and Optimization (pp. 865–879). Springer International Publishing. https://doi.org/10.1007/978-3-030-68154-8_74

Azizi, S., Mustafa, B., Ryan, F., Beaver, Z., Freyberg, J., Deaton, J., Loh, A., Karthikesalingam, A., Kornblith, S., Chen, T., Natarajan, V., & Norouzi, M. (2021). Big SelfSupervised Models Advance Medical Image Classification (arXiv:2101.05224). arXiv. https://doi.org/10.48550/arXiv.2101.05224

Cao, R., Wang, Y., Liang, Y., Gao, L., Zheng, J., Ren, J., & Wang, Z. (2022). Exploring the Impact of Negative Samples of Contrastive Learning: A Case Study of Sentence Embedding (arXiv:2202.13093). arXiv. https://doi.org/10.48550/arXiv.2202.13093

Caron, M., Misra, I., Mairal, J., Goyal, P., Bojanowski, P., & Joulin, A. (2021). Unsupervised Learning of Visual Features by Contrasting Cluster Assignments (arXiv:2006.09882). arXiv. https://doi.org/10.48550/arXiv.2006.09882

Chen, T., Kornblith, S., Norouzi, M., & Hinton, G. (2020). A Simple Framework for Contrastive Learning of Visual Representations (arXiv:2002.05709). arXiv. https://doi.org/10.48550/arXiv.2002.05709

Chen, X., Xie, S., & He, K. (2021). An Empirical Study of Training Self-Supervised Vision Transformers (arXiv:2104.02057). arXiv. https://doi.org/10.48550/arXiv.2104.02057

Chen, Y., Shen, X., Liu, Y., Tao, Q., & Suykens, J. A. K. (2023). Jigsaw-ViT: Learning Jigsaw Puzzles in Vision Transformer (arXiv:2207.11971). arXiv. https://doi.org/10.48550/arXiv.2207.11971

Chen, Z., Deng, Y., Wu, Y., Gu, Q., & Li, Y. (2022). Towards Understanding Mixture of Experts in Deep Learning (arXiv:2208.02813). arXiv. https://doi.org/10.48550/arXiv.2208.02813

Deng, J., Dong, W., Socher, R., Li, L.-J., Li, K., & Fei-Fei, L. (2009). ImageNet: A largescale hierarchical image database. 2009 IEEE Conference on Computer Vision and Pattern Recognition, 248–255. https://doi.org/10.1109/CVPR.2009.5206848

Devlin, J., Chang, M.-W., Lee, K., & Toutanova, K. (2019). BERT: Pre-training of Deep Bidirectional Transformers for Language Understanding (arXiv:1810.04805). arXiv. https://doi.org/10.48550/arXiv.1810.04805

Dosovitskiy, A., Beyer, L., Kolesnikov, A., Weissenborn, D., Zhai, X., Unterthiner, T., Dehghani, M., Minderer, M., Heigold, G., Gelly, S., Uszkoreit, J., & Houlsby, N. (2021). An Image is Worth
16x16 Words: Transformers for Image Recognition at Scale

(arXiv:2010.11929). arXiv. https://doi.org/10.48550/arXiv.2010.11929

Eslami, S., Meinel, C., & de Melo, G. (2023). PubMedCLIP: How Much Does CLIP

Benefit Visual Question Answering in the Medical Domain? In A. Vlachos & I. Augenstein (Eds.), Findings of the Association for Computational Linguistics: EACL 2023 (pp. 1181– 1193).

Association for Computational Linguistics. https://doi.org/10.18653/v1/2023.findings-eacl.88

Fürst, A., Rumetshofer, E., Lehner, J., Tran, V., Tang, F., Ramsauer, H., Kreil, D., Kopp, M., Klambauer, G., Bitto-Nemling, A., & Hochreiter, S. (2022). CLOOB: Modern Hopfield Networks with InfoLOOB Outperform CLIP (arXiv:2110.11316). arXiv. https://doi.org/10.48550/arXiv.2110.11316

Gao, Y., Wang, Z., Zheng, W.-S., Xie, C., & Zhou, Y. (2024). Sculpting Holistic 3D Representation in Contrastive Language-Image-3D Pre-training (arXiv:2311.01734). arXiv. https://doi.org/10.48550/arXiv.2311.01734

Goel, A., Tung, C., Lu, Y.-H., & Thiruvathukal, G. K. (2020). A Survey of Methods for LowPower Deep Learning and Computer Vision (arXiv:2003.11066). arXiv. https://doi.org/10.48550/arXiv.2003.11066

Gomez, R., Gomez, L., Gibert, J., & Karatzas, D. (2019). Self-Supervised Learning from

Web Data for Multimodal Retrieval (arXiv:1901.02004). arXiv. https://doi.org/10.48550/arXiv.1901.02004

Grill, J.-B., Strub, F., Altché, F., Tallec, C., Richemond, P. H., Buchatskaya, E., Doersch, C., Pires, B. A., Guo, Z. D., Azar, M. G., Piot, B., Kavukcuoglu, K., Munos, R., & Valko, M.

(2020). Bootstrap your own latent: A new approach to self-supervised Learning (arXiv:2006.07733). arXiv. https://doi.org/10.48550/arXiv.2006.07733

Gupta, A., Dollár, P., & Girshick, R. (2019). LVIS: A Dataset for Large Vocabulary Instance Segmentation (arXiv:1908.03195). arXiv. https://doi.org/10.48550/arXiv.1908.03195

Haralabopoulos, G., Torres, M. T., Anagnostopoulos, I., & McAuley, D. (2021). Text data augmentations: Permutation, antonyms and negation. Expert Systems with Applications,

177, 114769. https://doi.org/10.1016/j.eswa.2021.114769

He, K., Fan, H., Wu, Y., Xie, S., & Girshick, R. (2020). Momentum Contrast for Unsupervised Visual Representation Learning (arXiv:1911.05722). arXiv. https://doi.org/10.48550/arXiv.1911.05722

He, K., Zhang, X., Ren, S., & Sun, J. (2015). Deep Residual Learning for Image Recognition (arXiv:1512.03385). arXiv. https://doi.org/10.48550/arXiv.1512.03385

Huang, Y., Shakeri, F., Dolz, J., Boudiaf, M., Bahig, H., & Ayed, I. B. (2024). LP++: A Surprisingly Strong Linear Probe for Few-Shot CLIP (arXiv:2404.02285). arXiv. https://doi.org/10.48550/arXiv.2404.02285

Iscen, A., Caron, M., Fathi, A., & Schmid, C. (2024). Retrieval-Enhanced Contrastive VisionText Models (arXiv:2306.07196). arXiv. https://doi.org/10.48550/arXiv.2306.07196

Jia, C., Yang, Y., Xia, Y., Chen, Y.-T., Parekh, Z., Pham, H., Le, Q. V., Sung, Y., Li, Z., & Duerig, T. (2021). Scaling Up Visual and Vision-Language Representation Learning With Noisy Text Supervision (arXiv:2102.05918). arXiv. https://doi.org/10.48550/arXiv.2102.05918

Khan, A., Rauf, Z., Khan, A. R., Rathore, S., Khan, S. H., Shah, N. S., Farooq, U., Asif, H., Asif, A., Zahoora, U., Khalil, R. U., Qamar, S., Asif, U. H., Khan, F. B., Majid, A., & Gwak, J. (2023, December 1). A recent survey of vision transformers for medical image segmentation. arXiv.org. https://arxiv.org/abs/2312.00634

Khan, A., Rauf, Z., Sohail, A., Khan, A. R., Asif, H., Asif, A., & Farooq, U. (2023b). A survey of the vision transformers and their CNN-transformer based variants. Artificial Intelligence Review, 56(S3), 2917–2970. https://doi.org/10.1007/s10462-023-10595-0

Khan, A., Sohail, A., Fiaz, M., Hassan, M., Afridi, T. H., Marwat, S. U., Munir, F., Ali, S., Naseem, H., Zaheer, M. Z., Ali, K., Sultana, T., Tanoli, Z., & Akhter, N. (2024, August 30). A survey of the Self Supervised Learning Mechanisms for vision Transformers. arXiv.org. https://arxiv.org/abs/2408.17059

Khan, A., Sohail, A., Zahoora, U., & Qureshi, A. S. (2020). A survey of the recent architectures of deep convolutional neural networks. Artificial Intelligence Review, 53(8), 5455–5516. https://doi.org/10.1007/s10462-020-09825-6

Rauf, Z., Khan, A. R., & Khan, A. (2024, July 27). Channel boosted CNN-Transformer-based Multi-Level and Multi-Scale nuclei segmentation. arXiv.org. https://arxiv.org/abs/2407.19186

Khosla, P., Teterwak, P., Wang, C., Sarna, A., Tian, Y., Isola, P., Maschinot, A., Liu, C., & Krishnan, D. (2021). Supervised Contrastive Learning (arXiv:2004.11362). arXiv. https://doi.org/10.48550/arXiv.2004.11362

Krishna, T., McGuinness, K., & O'Connor, N. (2021). Evaluating Contrastive Models for Instance-based Image Retrieval. Proceedings of the 2021 International Conference on Multimedia Retrieval, 471–475. https://doi.org/10.1145/3460426.3463585

Kwon, G., Cai, Z., Ravichandran, A., Bas, E., Bhotika, R., & Soatto, S. (2023). Masked Vision and Language Modeling for Multi-modal Representation Learning (arXiv:2208.02131). arXiv. https://doi.org/10.48550/arXiv.2208.02131

Lan, Z., Chen, M., Goodman, S., Gimpel, K., Sharma, P., & Soricut, R. (2020). ALBERT: A Lite BERT for Self-supervised Learning of Language Representations (arXiv:1909.11942). arXiv. https://doi.org/10.48550/arXiv.1909.11942

Liao, Y., Jiang, X., & Liu, Q. (2020). Probabilistically Masked Language Model Capable of Autoregressive Generation in Arbitrary Word Order. In D. Jurafsky, J. Chai, N. Schluter, & J. Tetreault (Eds.), Proceedings of the 58th Annual Meeting of the Association for Computational Linguistics (pp. 263–274). Association for Computational Linguistics. https://doi.org/10.18653/v1/2020.acl-main.24

Lin, T.-Y., Maire, M., Belongie, S., Hays, J., Perona, P., Ramanan, D., Dollár, P., & Zitnick, C. L. (2014). Microsoft COCO: Common Objects in Context. In D. Fleet, T. Pajdla, B. Schiele, & T. Tuytelaars (Eds.), Computer Vision – ECCV 2014 (pp. 740–755). Springer International Publishing. https://doi.org/10.1007/978-3-319-10602-1_48

Lu, Z. (2022). Brief Introduction to Contrastive Learning Pretext Tasks for Visual Representation (arXiv:2210.03163). arXiv. https://doi.org/10.48550/arXiv.2210.03163

Luo, Z., Xi, Y., Zhang, R., Li, G., Zhao, Z., & Ma, J. (2022). Conditioned Masked Language and Image Modeling for Image-Text Dense Retrieval. In Y. Goldberg, Z. Kozareva, & Y. Zhang (Eds.), Findings of the Association for Computational Linguistics: EMNLP 2022 (pp. 130–140). Association for Computational Linguistics. https://doi.org/10.18653/v1/2022.findings-emnlp.10

Masked Image Contrastive Learning for Efficient Visual Conceptual Pre-training. (n.d.). Retrieved March 7, 2025, from https://arxiv.org/html/2411.09858v1

Misra, I., & Maaten, L. van der. (2019). Self-Supervised Learning of Pretext-Invariant Representations (arXiv:1912.01991). arXiv. https://doi.org/10.48550/arXiv.1912.01991

Mu, N., Kirillov, A., Wagner, D., & Xie, S. (2021). SLIP: Self-supervision meets LanguageImage Pre-training (arXiv:2112.12750). arXiv. https://doi.org/10.48550/arXiv.2112.12750

Mustafa, B., Riquelme, C., Puigcerver, J., Jenatton, R., & Houlsby, N. (2022). Multimodal Contrastive Learning with LIMoE: The Language-Image Mixture of Experts

(arXiv:2206.02770). arXiv. https://doi.org/10.48550/arXiv.2206.02770

Noroozi, M., & Favaro, P. (2017). Unsupervised Learning of Visual Representations by Solving Jigsaw Puzzles (arXiv:1603.09246). arXiv. https://doi.org/10.48550/arXiv.1603.09246

Novack, Z., McAuley, J., Lipton, Z. C., & Garg, S. (2023). CHiLS: Zero-Shot Image Classification with Hierarchical Label Sets (arXiv:2302.02551). arXiv. https://doi.org/10.48550/arXiv.2302.02551

Oord, A. van den, Li, Y., & Vinyals, O. (2019). Representation Learning with Contrastive Predictive Coding (arXiv:1807.03748). arXiv. https://doi.org/10.48550/arXiv.1807.03748

Pandey, A., & Vishwakarma, D. K. (2024). Contrastive Learning-based Multi Modal Architecture for Emoticon Prediction by Employing Image-Text Pairs

(arXiv:2408.02571). arXiv. https://doi.org/10.48550/arXiv.2408.02571

Peng, D., & Li, Z. (2024). Robust visual question answering via polarity enhancement and contrast. Neural Networks: The Official Journal of the International Neural Network Society, 179, 106560. https://doi.org/10.1016/j.neunet.2024.106560

Philbin, J., Chum, O., Isard, M., Sivic, J., & Zisserman, A. (2007). Object retrieval with large vocabularies and fast spatial matching. 2007 IEEE Conference on Computer Vision and Pattern Recognition, 1–8. https://doi.org/10.1109/CVPR.2007.383172

Philbin, J., Chum, O., Isard, M., Sivic, J., & Zisserman, A. (2008). Lost in quantization:

Improving particular object retrieval in large scale image databases. 2008 IEEE Conference on Computer Vision and Pattern Recognition, 1–8. https://doi.org/10.1109/CVPR.2008.4587635

Pourpanah, F., Abdar, M., Luo, Y., Zhou, X., Wang, R., Lim, C. P., Wang, X.-Z., & Wu, Q. M.

J. (2022). A Review of Generalized Zero-Shot Learning Methods. IEEE Transactions on Pattern Analysis and Machine Intelligence, 1–20. https://doi.org/10.1109/TPAMI.2022.3191696

Qian, R., Meng, T., Gong, B., Yang, M.-H., Wang, H., Belongie, S., & Cui, Y. (2021). Spatiotemporal Contrastive Video Representation Learning (arXiv:2008.03800). arXiv. https://doi.org/10.48550/arXiv.2008.03800

Radford, A., Kim, J. W., Hallacy, C., Ramesh, A., Goh, G., Agarwal, S., Sastry, G., Askell, A., Mishkin, P., Clark, J., Krueger, G., & Sutskever, I. (2021). Learning Transferable Visual Models From Natural Language Supervision (arXiv:2103.00020). arXiv. https://doi.org/10.48550/arXiv.2103.00020

Ruslim, A. R., Yudistira, N., & Setiawan, B. D. (2023). Mixture of Self-Supervised Learning (arXiv:2307.14897). arXiv. https://doi.org/10.48550/arXiv.2307.14897

Sermanet, P., Lynch, C., Chebotar, Y., Hsu, J., Jang, E., Schaal, S., & Levine, S. (2018). Time- Contrastive Networks: Self-Supervised Learning from Video (arXiv:1704.06888). arXiv. https://doi.org/10.48550/arXiv.1704.06888

Singh, H., Zhang, P., Wang, Q., Wang, M., Xiong, W., Du, J., & Chen, Y. (2023). Coarse-toFine Contrastive Learning in Image-Text-Graph Space for Improved Vision-Language Compositionality. In H. Bouamor, J. Pino, & K. Bali (Eds.), Proceedings of the 2023 Conference on Empirical Methods in Natural Language Processing (pp. 869–893). Association for Computational Linguistics. https://doi.org/10.18653/v1/2023.emnlp-main.56

Sonawale, R., & Shaikh, A. (n.d.). Visual Mind: Visual Question Answering (VQA) with CLIP Model. Engineering Technology, 12.

Tosh, C., Krishnamurthy, A., & Hsu, D. (2021). Contrastive learning, multi-view redundancy, and linear models (arXiv:2008.10150). arXiv. https://doi.org/10.48550/arXiv.2008.10150

Uy, M. A., Pham, Q.-H., Hua, B.-S., Nguyen, D. T., & Yeung, S.-K. (2019). Revisiting Point Cloud Classification: A New Benchmark Dataset and Classification Model on Real-World Data (arXiv:1908.04616). arXiv. https://doi.org/10.48550/arXiv.1908.04616

Wang, J., Chen, Y., & Yu, S. X. (2024). Pose-Aware Self-Supervised Learning with Viewpoint Trajectory Regularization (arXiv:2403.14973; Version 2). arXiv. https://doi.org/10.48550/arXiv.2403.14973

Wang, P., Li, Q., Yu, L., Wang, Z., Li, A., & Jin, H. (2024). Moderator: Moderating TexttoImage Diffusion Models through Fine-grained Context-based Policies. Proceedings of the 2024 on ACM SIGSAC Conference on Computer and Communications Security, 1181–1195. https://doi.org/10.1145/3658644.3690327

Wen, K., Tan, Z., Cheng, Q., Chen, C., & Gu, X. (2022). Contrastive Cross-Modal Knowledge Sharing Pre-training for Vision-Language Representation Learning and Retrieval (arXiv:2207.00733). arXiv. https://doi.org/10.48550/arXiv.2207.00733

Xie, C.-W., Sun, S., Xiong, X., Zheng, Y., Zhao, D., & Zhou, J. (2023). RA-CLIP: Retrieval Augmented Contrastive Language-Image Pre-Training. 2023 IEEE/CVF Conference on Computer Vision and Pattern Recognition (CVPR), 19265–19274. https://doi.org/10.1109/CVPR52729.2023.01846

Xu, H., Ghosh, G., Huang, P.-Y., Okhonko, D., Aghajanyan, A., Metze, F., Zettlemoyer, L., & Feichtenhofer, C. (2021). VideoCLIP: Contrastive Pre-training for Zero-shot Video-Text Understanding (arXiv:2109.14084). arXiv. https://doi.org/10.48550/arXiv.2109.14084 Yan, S., Zhu, T., Wang, Z., Cao, Y., Zhang, M., Ghosh, S., Wu, Y., & Yu, J. (2023).

VideoCoCa: Video-Text Modeling with Zero-Shot Transfer from Contrastive Captioners

(arXiv:2212.04979). arXiv. https://doi.org/10.48550/arXiv.2212.04979

Yang, Z., Du, H., Niyato, D., Wang, X., Zhou, Y., Feng, L., Zhou, F., Li, W., & Qiu, X. (2024).

Revolutionizing Wireless Networks with Self-Supervised Learning: A Pathway to Intelligent Communications (arXiv:2406.06872). arXiv. https://doi.org/10.48550/arXiv.2406.06872

Yu, W., Zhu, C., Zhao, T., Guo, Z., & Jiang, M. (2021). Sentence-Permuted Paragraph Generation. In M.-F. Moens, X. Huang, L. Specia, & S. W. Yih (Eds.), Proceedings of the 2021 Conference on Empirical Methods in Natural Language Processing (pp. 5051–5062).

Association for Computational Linguistics. https://doi.org/10.18653/v1/2021.emnlp-main.412

Yuchen Guo, G. D. (n.d.). Zero-Shot Learning With attribute Selection. AAAI. Retrieved March

7, 2025, from https://aaai.org/papers/12251-zero-shot-learning-with-attribute-selection/

Zheng, M., Wang, F., You, S., Qian, C., Zhang, C., Wang, X., & Xu, C. (2021). Weakly Supervised Contrastive Learning (arXiv:2110.04770). arXiv. https://doi.org/10.48550/arXiv.2110.04770

Zhong, Y., Tang, H., Chen, J., Peng, J., & Wang, Y.-X. (2022). Is Self-Supervised Learning More Robust Than Supervised Learning? (arXiv:2206.05259). arXiv.

https://doi.org/10.48550/arXiv.2206.05259

Giakoumoglou, N., & Stathaki, T. (2025). SynCo: Synthetic Hard Negatives for Contrastive Visual Representation Learning. arXiv. https://arxiv.org/abs/2410.2401

Luthra, A., Yang, T., & Galanti, T. (2025). Self-supervised contrastive learning is approximately supervised contrastive learning. arXiv. https://arXiv.org/abs/2506.04411